\documentclass{article}
\usepackage{ijcai24}

\usepackage{times}
\usepackage{soul}
\usepackage{url}
\usepackage[hidelinks]{hyperref}
\usepackage[utf8]{inputenc}
\usepackage[small]{caption}
\usepackage{graphicx}
\usepackage{amsmath}
\usepackage{amsthm}
\usepackage{booktabs}
\usepackage{algorithm}
\usepackage{algorithmic}
\usepackage[switch]{lineno}
\usepackage[utf8]{inputenc} 
\usepackage[T1]{fontenc}    
\usepackage{hyperref}       
\usepackage{url}            
\usepackage{booktabs}       
\usepackage{amsfonts}       
\usepackage{nicefrac}       
\usepackage{microtype}      
\usepackage{xcolor}         
\usepackage{wrapfig}
\usepackage{times}
\usepackage{soul}
\usepackage{amssymb}
\usepackage{mathtools}
\usepackage{subfiles}
\usepackage{amssymb}
\usepackage{mathtools}
\usepackage{subcaption}
\usepackage{multirow}
\usepackage{xspace}
\usepackage{makecell}
\usepackage{verbatim}
\usepackage{array}
\usepackage{placeins}
\usepackage{stfloats}
\usepackage{enumitem}
\usepackage{color}

\newcommand{\methodName}{\textsc{EAB-FL}\xspace}
\newcommand{\fedavg}{\textsc{FedAvg}\xspace}
\newcommand{\fedfb}{\textsc{FedFB}\xspace}
\newcommand{\fairfed}{\textsc{FairFed}\xspace}
\usepackage{multirow}
\usepackage{xspace}
\usepackage{makecell}
\usepackage{verbatim}
\usepackage{array}
\usepackage{placeins}
\usepackage{stfloats}
\usepackage{color}
\newcommand{\qffl}{\textsc{q-FFL}\xspace}
\newcommand{\gifair}{\textsc{GIFAIR-FL}\xspace}
\title{\methodName: Exacerbating Algorithmic Bias through Model Poisoning Attacks \\in Federated Learning}
\author{
Syed Irfan Ali Meerza
\And
Jian Liu\\
\affiliations
University of Tennessee, Knoxville\\
\emails
smeerza@vols.utk.edu,
jliu@utk.edu,
}
\begin{document}

\maketitle

\begin{abstract}
    Federated Learning (FL) is a technique that allows multiple parties to train a shared model collaboratively without disclosing their private data. It has become increasingly popular due to its distinct privacy advantages. However, FL models can suffer from biases against certain demographic groups (e.g., racial and gender groups) due to the heterogeneity of data and party selection. Researchers have proposed various strategies for characterizing the group fairness of FL algorithms to address this issue. However, the effectiveness of these strategies in the face of deliberate adversarial attacks has not been fully explored. Although existing studies have revealed various threats (e.g., model poisoning attacks) against FL systems caused by malicious participants, their primary aim is to decrease model accuracy, while the potential of leveraging poisonous model updates to exacerbate model unfairness remains unexplored. In this paper, we propose a new type of model poisoning attack, \methodName, with a focus on exacerbating group unfairness while maintaining a good level of model utility. Extensive experiments on three datasets demonstrate the effectiveness and efficiency of our attack, even with state-of-the-art fairness optimization algorithms and secure aggregation rules employed. Code is available at \url{https://github.com/irfanMee/EAB-FL}
\end{abstract}

\vspace{-3mm}
\section{Introduction}
\vspace{-1mm}

Federated Learning (FL)~\cite{konevcny2016federated} has recently emerged as a promising solution that enables multiple clients to collaboratively learn a shared prediction model while keeping all the training data on the device. Due to its privacy-preserving nature, in recent years, FL has benefited a wide variety of privacy-sensitive application domains, such as medical research~\cite{rauniyar2023federated}, and financial fraud~\cite{liu2023efficient}, etc. However, the distributed nature of FL makes it inherently vulnerable to poisoning attacks, in which the model can be compromised by malicious clients uploading malicious model updates. Without a central authority to validate clients' participation, these malicious clients can indirectly and consequently manipulate the parameters of the learned model and thereby reduce its overall performance~\cite{cao2022mpaf}.

In addition, compared to centralized learning, FL models are more susceptible to algorithmic bias against specific demographic groups (e.g., racial and gender groups) due to its inherent characteristics, such as data heterogeneity, party selection, and client dropping out~\cite{abay2020mitigating}. Compounding this issue, the involvement of malicious participants within FL environments can further exacerbate these biases.

\textbf{Attacker's Motivations.}
Attacking FL models through poisoning attacks to exacerbate their unfairness could provide an attacker with various benefits. For instance, e-commerce websites can be targeted with fairness attacks on their recommendation algorithms to suggest certain groups of products or services for the benefit of their providers while harming others. In the case of loan applications, an attacker might attempt to manipulate the FL model to unfairly discriminate against or favor specific groups of people, resulting in unjust loan decisions. Furthermore, the attacker can also attack the models used in criminal justice areas to make them unfair to reduce the trust and credibility in the criminal justice system. Given the aforementioned potential and motivations of fairness attacks, it is crucial to have a comprehensive understanding of the attack surfaces of FL, particularly in the domain of fairness.

\textbf{Existing Attacks.} There are a few studies exploring poisoning attacks against FL systems, either by adding poisonous instances or adversarially changing model updates~\cite{cao2022mpaf,bagdasaryan2020backdoor}. However, these attacks are proposed with the purpose of reducing the model's classification accuracy without any regard for the model's fairness. In centralized machine learning, there have been attacking efforts to exacerbate algorithmic bias, such as the gradient-based poisoning attacks~\cite{solans2021poisoning} and the anchoring and influence attacks~\cite{mehrabi2021exacerbating}, which aim to maximize the covariance between the sensitive attributes and the decision outcome to affect the model fairness. However, these attacks can hardly be adapted to the FL settings due to the challenges in measuring the impact of each data sample on fairness violation in the learned model. This is mainly because the data used in FL is often decentralized and not directly accessible, making it hard to evaluate the specific impact of each sample on the model. To the best of our knowledge, the potential of leveraging poisoning attacks in FL to exacerbate group unfairness remains unexplored.

\textbf{Our Attack.} In this paper, we design a new type of model poisoning attack, \methodName, where an adversary can introduce or exacerbate algorithmic bias against certain groups of individuals or samples while maintaining a relatively good level of model utility (i.e., classification accuracy). We assume the adversary has compromised a small fraction of client devices (a.k.a., malicious clients) and can manipulate the training process on these devices. To maintain the overall model utility, \methodName first identifies the redundant space of the locally trained model by applying layer-wise relevance propagation (LRP)~\cite{bach2015pixel}. This space comprises neurons that remain relatively stable during the learning process and exhibit minimal correlation with the prediction task for the privileged demographic group (i.e., the one that the adversary does not want to impact). Subsequently, \methodName adjusts the model parameters within the redundant space by solving an optimization problem on a subset of local datasets that adversely influences the model's performance for the targeted group. Consequently, our method can preserve high utility for the privileged group while simultaneously reducing it for the targeted group, thus inducing algorithmic bias in the model.

To make the attack less suspicious and more generalized across different FL settings, \methodName~has the following major properties: \textit{(1) High Model Utility:} Unlike other poisoning attacks in FL that target model classification accuracy, our attack aims to exacerbate group unfairness only, with a minimum impact on the model's overall utility, thereby rendering the attack less noticeable; \textit{(2) Persistence and Stealth Attack:} We intent to embed adversarial features into the redundant space of the model to improve the persistence of attack while remaining the alterations of the model parameters minimal, aiming to ensure that the attack remains robust against secure aggregations; and \textit{(3) Effective under Fairness Optimizations:} Our attack can remain effective under existing fairness optimization strategies (e.g., \fedfb~\cite{zeng2021improving}, \fairfed~\cite{ezzeldin2021fairfed}), as the adversary optimizes the poisoning each time it participates in the training, which allows it to nullify the effect imposed by any fairness optimization methods.

Through extensive experiments on three datasets with different fairness optimization, we demonstrate the effectiveness of~\methodName in achieving the desired goal of exacerbating group unfairness. We also evaluate our attack under various state-of-the-art secure aggregation rules in FL, and the results demonstrate its sustained efficacy.

\vspace{-3mm}
\section{Related Work}
\vspace{-1mm}

Model poisoning attacks against FL systems have received considerable attention in recent years. These attacks are designed to manipulate the global model by tampering with local training processes on a fraction of participating clients. For instance, malicious clients can significantly degrade the performance of the global model by adding random noise to the local model to mislead the global model~\cite{hossain2021desmp,cao2022mpaf}. Xingchen \textit{et al.}~\cite{zhou2021deep} proposed an optimization-based model poisoning attack to inject poisonous neurons into the model’s redundant space, identified using the Hessian matrix. 
Moreover, it has been shown that the adversary can replace the aggregated model with the malicious model through one compromised device to perform model-replacement attacks~\cite{xie2020dba,bagdasaryan2020backdoor}. Henger \textit{et al.}~\cite{li2022learning} proposed a model poisoning attack using reinforcement learning, where malicious clients collectively learn the data distribution to launch an optimal attack.
While the aforementioned attacks have shown a great chance of compromising FL models, they only target the model's utility by either decreasing its overall classification accuracy or making specific test samples classified as adversary-desired labels. 

In addition, it has been shown that FL models are more likely to suffer from algorithmic bias compared to centralized learning due to their inherent characteristics, such as data heterogeneity, party selection, and client dropping out~\cite{abay2020mitigating}. To address this issue, initial studies~\cite{li2019fair,mohri2019agnostic,lyu2020collaborative,li2021ditto,zhang2020hierarchically} focus on client parity and aim to promote equalized accuracies across all participating clients in FL. This is typically achieved by reducing the client's performance disparity~\cite{li2019fair} or maximizing the performance of the worst client~\cite{mohri2019agnostic}. More recent studies have addressed group fairness~\cite{abay2020mitigating,zhang2020fairfl,rodriguez2021enforcing,chu2021fedfair,ezzeldin2021fairfed,du2021fairness}. These works propose solutions for providing fair performance across different sensitive groups. These studies mainly utilize deep multi-agent reinforcement learning~\cite{zhang2020fairfl}, re-weighting mechanisms~\cite{abay2020mitigating,ezzeldin2021fairfed,du2021fairness}, optimization with fairness constraints (e.g., via alternating gradient projection~\cite{chu2021fedfair} or the modified method of differential multipliers~\cite{rodriguez2021enforcing}) to achieve group fairness. 

In adversarial scenarios, attacks targeting fairness measures in machine learning are a relatively new concept, and only a few studies have been proposed in this area. Solans et al.~\cite{solans2021poisoning} were among the first to propose a fairness attack that uses a gradient-based poisoning attack to introduce classification disparities among different groups. Mehrabi et al.~\cite{mehrabi2021exacerbating} proposed anchoring and influence attacks to introduce algorithmic bias in machine learning algorithms. More recently, Chhabra et al.~\cite{chhabra2022robust} proposed a black-box fairness attack on clustering algorithms. However, all these attacks have been proposed in centralized learning settings. To the best of our knowledge, the potential of adversarial attacks aiming to exacerbate model unfairness in FL remains unexplored, which is vital for us to fully understand the attack surfaces of FL and thereby help facilitate corresponding mitigations to improve its resilience.

\vspace{-3mm}
\section{Preliminaries}
\vspace{-1mm}
\subsection{Federated Learning}
\vspace{-1mm}
Federated Learning (FL) is a collaborative approach in machine learning where a global model is trained across multiple distributed clients under the supervision of a central server. This method is distinct because it does not require direct access to client data, thereby enhancing privacy and data security. FL's primary aim is to optimize the global model's parameters while effectively utilizing the diverse, decentralized data held by each client. The key formula governing FL is:
{\footnotesize
\begin{equation} 
\setlength{\abovedisplayskip}{3pt}
\setlength{\belowdisplayskip}{3pt}
  \underset{\theta_{g}}{\min} \quad f(\theta_{g}) = \sum_{k=1}^{n} p_{k}\mathcal{L}_{k}(\theta_{g}), \quad \mathcal{L}_{k} = \frac{1}{d_{k}} \sum_{j_{k}=1}^{d_{k}} l_{j_{k}}(\theta_{g}),
\label{eq1}
\end{equation}
}
where, \( \theta_{g} \) represents the global model parameters, \( n \) is the total number of clients, \( p_{k} \) is the probability of each client \( k \)'s participation, \( \mathcal{L}_{k} \) is the empirical loss for client \( k \), \( l_{j_{k}} \) is the loss for each data sample \( j \) of client \( k \), and \( d_{k} \) denotes the number of data samples of client \( k \). The optimization in FL typically involves selecting a subset of clients in each training round, based on their participation probability, and then applying local optimizers like Stochastic Gradient Descent (SGD). A widely used model aggregation method, FedAvg~\cite{mcmahan2017communication}, involves averaging the participating client models. However, this approach often results in performance inconsistencies among clients, with potential biases towards clients with larger datasets or those participating more frequently. This can inadvertently introduce biases against certain demographic groups in the dataset.

\subsection{Group Fairness in FL}
\vspace{-1mm}
In a binary classification task, we deal with training samples of the form $(x_{1}, y_{1}, g_{1}), ..., (x_{d}, y_{d}, g_{d})$ where each example consists of an instance $x_i \in X$, a label $y_i \in Y$, and a sensitive attribute $g_{i} \in G$. The goal is to develop a classification model, denoted as \(f(\theta): X \rightarrow Y\), which aims to minimize the cumulative loss, $\mathcal{L}_m(\theta, \mathcal{D}) = \sum_{(x,y)\in \mathcal{D}} l(f(x,\theta), y)$, over the training dataset $\mathcal{D}={(X,Y,G)}$ to find the optimal parameters. In the context of Federated Learning (FL), the framework strives not only for accuracy but also for fairness in model predictions concerning the sensitive attribute $g_i$. Fairness is evaluated based on certain notions, such as demographic parity and equal opportunity. A model is considered fair from a group fairness perspective if it performs equally well for both the privileged group ($g_i = 1$) and the underprivileged group ($g_i = 0$). For a model yielding binary predictions $\hat{Y}$, given data samples $X$ and their corresponding labels $Y$, we use the following two metrics to assess the model's fairness: 

\noindent \textbf{Demographic Parity~\cite{dwork2012fairness}:} If a classifier's predictions $\hat{Y}$ is statistically independent of the sensitive characteristic $G$, it meets demographic parity under a distribution $(X, Y, G)$.
This is equivalent to $\mathbb{E}[\hat{Y}|G=a] = \mathbb{E}[\hat{Y}]$, where $a=0$ or $1$ for a binary group.
Demographic parity can be defined as: 
{\footnotesize
\begin{equation} 
\setlength{\abovedisplayskip}{3pt}
\setlength{\belowdisplayskip}{3pt}
    Pr\{\hat{Y}=1|G=1\} = Pr\{\hat{Y}=1|G=0\}.
\label{eq2}
\vspace{-0mm}
\end{equation}
}
\noindent \textbf{Equal Opportunity~\cite{hardt2016equality}:} If a classifier's predictions $\hat{Y}$ is conditionally independent of the sensitive feature given the label, it meets equalized opportunity under a distribution $(X, Y, G)$. This is the same as $\mathbb{E}[\hat{Y}|G=a, Y=1] = \mathbb{E}[\hat{Y}|Y=1]$. In this case, we want the true positive rate $Pr\{\hat{Y}=1|Y=1\}$ to be the same for each population with no regard for the errors when $Y=0$. Equal Opportunity thus can be defined as:
{\footnotesize
\begin{equation} 
\setlength{\abovedisplayskip}{3pt}
\setlength{\belowdisplayskip}{3pt}
    Pr\{\hat{Y}=1|Y=1,G=1\} = Pr\{\hat{Y}=1|Y=1,G=0\}.
\label{fig-1}
\vspace{-1mm}
\end{equation}
}

\subsection{Influence Score}
In~\methodName, to exacerbate model bias, each malicious client needs to identify a subset of their local training samples that can detrimentally affect the performance of a specific demographic group (i.e., targeted group). To quantify the impact of each local training sample on the model’s performance concerning the targeted
demographic group, we use ``influence score''~\cite{wang2022understanding} to assess how a model's prediction on the data samples from the targeted group would change if a training sample $(x_i,y_i,g_i)$ is excluded from the training dataset, particularly under fairness constraints imposed on the classifier. 
Specifically, the influence score of a training example $(x_i,y_i,g_i)$ concerning a specific demographic group (e.g., $g_j = \tau$) can be represented as:
{\footnotesize\begin{equation}
\setlength{\abovedisplayskip}{3pt}
\setlength{\belowdisplayskip}{3pt}
\begin{split}
     \text{inf}_l(\mathcal{D}, i)\approx \int_{g_{j} = \tau} \Theta(x_i, x_j ; \theta) (\frac{\partial \mathcal{L}(w, y_i)}{\partial w}\bigg|_{w = f(x_i;\theta)} \\
     + \frac{\partial \phi(f,g_i)}{\partial l})\bigg|_{f(x_i;\theta)}) \text{d}Pr(x_j,y_j,g_j),
\end{split}
\label{eq:influ}
\end{equation}}
where $\text{inf}_l(\mathcal{D}, i) \in \mathbb{R}$, $\Theta$ is the Neural Tangent Kernel (NTK)~\cite{jacot2018neural} and $\theta$ represents the parameters of the classification model. $\phi(f,g_i)$ denotes a differentiable surrogate for commonly used fairness constraints, such as Demographic Parity or Equal Opportunity, $\mu$ is a tolerance parameter for fairness deviations, and $\text{d}Pr(x_j, y_j, z_j)$ refers to the differential probability measure over the data distribution $\mathcal{D}$.
We solve this equation by calculating the Jacobian of the function that multiplies the model's output with the demographics attribute $(g_j = \tau)$, and then compute a kernel matrix from the gradients.
A positive value suggests that including the training sample $(x_i,y_i,g_i)$ improves the model's performance for the targeted demographic group, while a negative value implies it hinders accuracy. The magnitude of the value shows the strength of this impact.


\begin{figure}[t]
  \centering
  \includegraphics[width=0.60\linewidth]{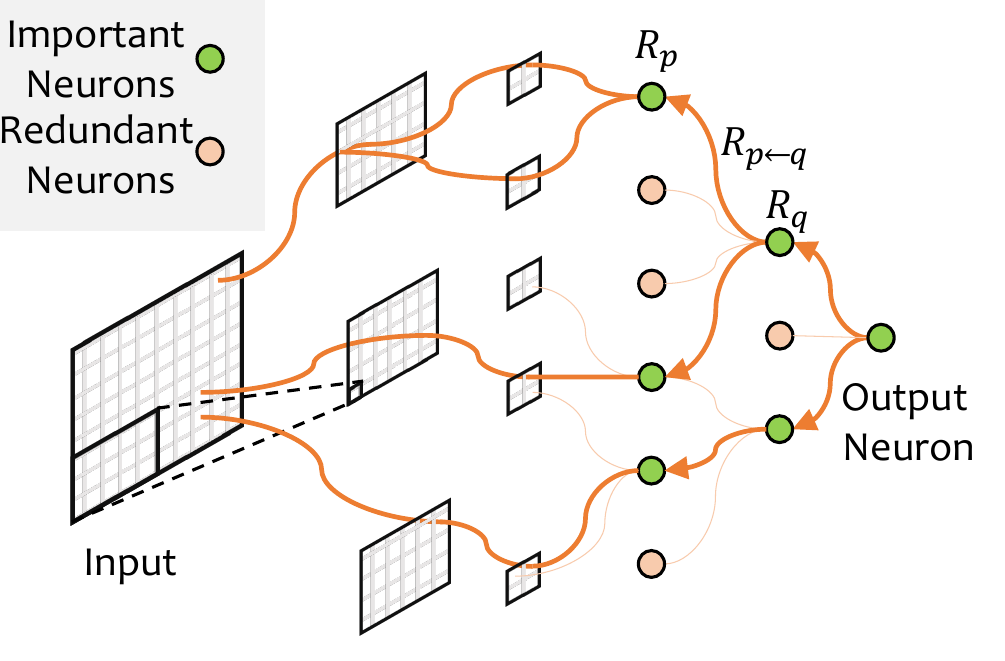}
  \vspace{-2mm}
  \caption{Illustration of the LRP procedure.}
  \label{fig:LRP}
  \vspace{-2mm}
\end{figure}

\subsection{Layer-wise Relevance Propagation}
\label{subsec:lrp}
\vspace{-1mm}
Layer-wise Relevance Propagation (LRP)~\cite{bach2015pixel} is an explanation technique, aiming to propagate the model prediction $f(x,\theta)$ backward in the model to quantify the contributions of each neuron to the model prediction. Specifically, during the backward propagation of LRP, as shown in Figure~\ref{fig:LRP}, each neuron redistributes the received relevance scores to the preceding layer in equal proportions.
By examining the propagated relevance scores, LRP can determine the degree of influence each neuron has on the model prediction. Neurons with higher relevance scores are deemed more ``important'' in decision-making, while those with lower scores are considered relatively ``redundant''. If we consider $p$ and $q$ as neurons in two successive layers, the relevance scores $(R_q)_p$  at one layer are transferred to the neurons in the layer below by applying the following rule:
\begin{equation}
\setlength{\abovedisplayskip}{3pt}
\setlength{\belowdisplayskip}{3pt}
    R_p = \sum_{q} \frac{z_{pq}}{\sum_{p}z_{pq}}R_q,
    \label{eq:lrp}
\end{equation}
where $z_{pq}$ is the product of the activation of neuron $p$ and the weight of the connection from neuron $p$ to neuron $q$, which quantifies the contribution of neuron $p$ to the relevance of neuron $q$. 
The denominator acts as a normalizing factor to uphold the conservation principle, ensuring that the sum of relevance scores propagated from a neuron matches the sum of scores it received. In \methodName, we use LRP procedure to identify the model's redundant space and confine all poisoning alterations within this space to exacerbate model bias while ensuring minimal impact on the model utility.




\vspace{-3mm}
\section{Poisoning Attacks against Fairness in FL}
\vspace{-1mm}
\subsection{Threat Model and Adversarial Goals}

\textbf{Threat Model:} 
In this work, we assume an adversary has compromised a small fraction of client devices (a.k.a., malicious clients), and the adversary can tamper with the local training process on the compromised client devices during the learning to exacerbate model bias. The attack does not involve manipulating the local datasets, making it easy to bypass any security measures focused on data integrity. 
During the FL participation process, we assume that the central server may employ security measures to validate the credibility of the submitted model updates~\cite{pillutla2022robust,bhagoji2019analyzing}. For instance, the server can evaluate the accuracy of the submitted model updates on a validation set, and the server can also verify model updates through secure aggregation rules (e.g., Krum~\cite{yin2018byzantine}) to mitigate or reject anomalous model updates.

\textbf{Adversarial Goal:} Different from traditional poisoning attacks, which only aim to decrease the model's utility, the adversary in our attack aims to rely on the compromised client devices to exacerbate the algorithmic bias in the learned global model while maintaining a relatively good level of model utility (e.g., classification accuracy).

\begin{figure}[t]
  \centering
  \includegraphics[width=0.99\linewidth]{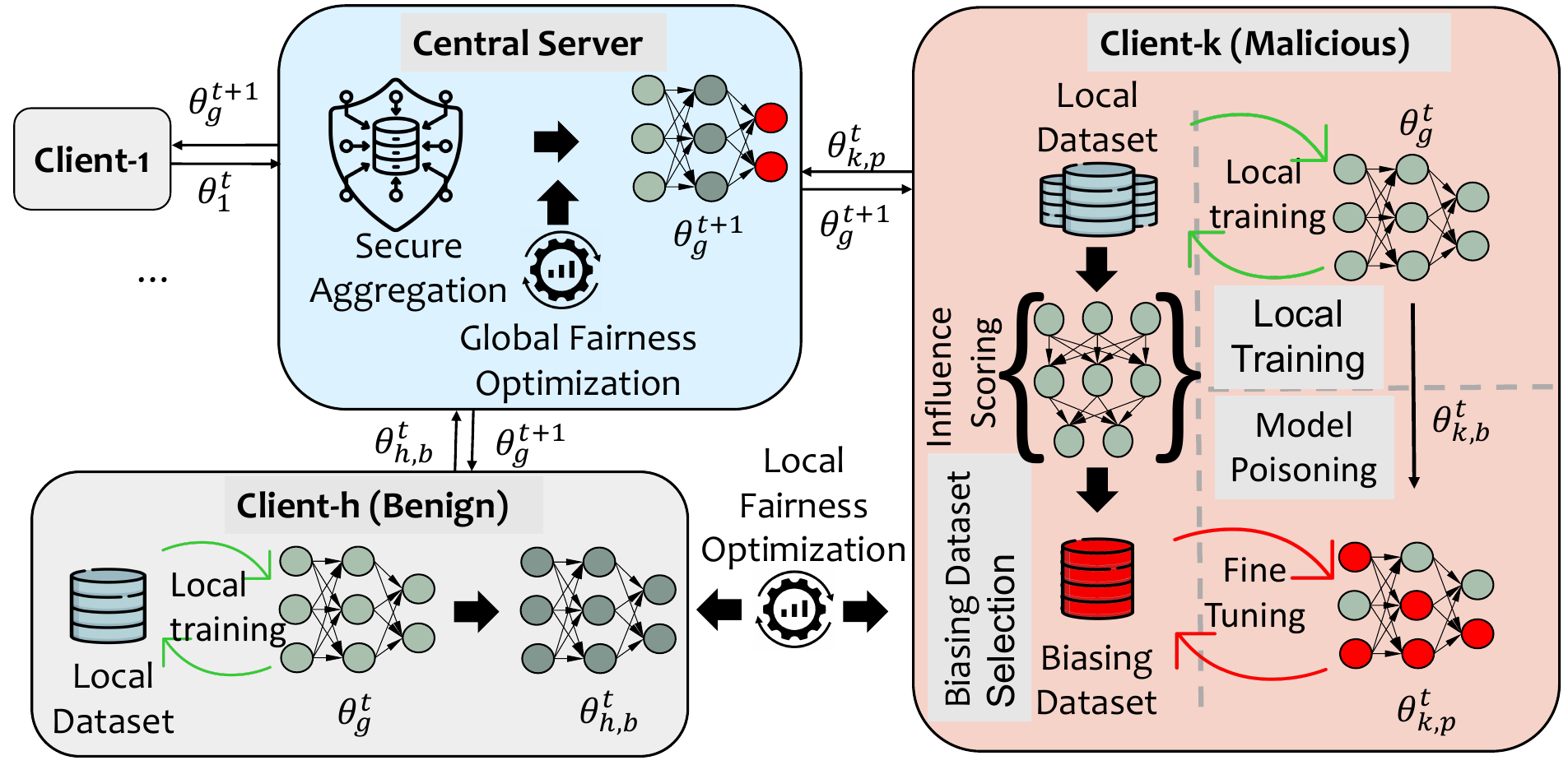}
  \caption{Illustration of the proposed \methodName. \scriptsize text before after}
  \label{fig:system_overview}
  \vspace{-2mm}
\end{figure}

\textbf{Adversary's Knowledge:} 
In the threat model, we assume that the adversary has full knowledge of the structure and parameters of the global model $\theta_g^t$ received at each communication round. The adversary is considered to be active, meaning that the adversary can tamper with the local training process and locally deploy an optimization-based approach on each compromised client device. However, the adversary does not have any knowledge of the aggregation rule used on the server or the fairness optimization methods applied in FL. We assume a non-colluding malicious client, where the adversary has no partner to exchange any information nor does it have any knowledge of the benign clients’ local training.

\vspace{-1mm}
\subsection{Attack Design Overview}
\vspace{-1mm}
In this work, we propose an optimization-based model poisoning attack in FL, \methodName, that targets the group fairness measure of the learned global model. As shown in Figure~\ref{fig:system_overview}, the FL system consists of a central server and $n$ clients, a small fraction of which have been compromised by an adversary. During each communication round, benign (non-compromised) clients (e.g., the client $h$) train the global model ($\theta_g^t$) provided by the central server on their local datasets for a certain number of epochs and send the updated model ($\theta_{h,b}^t$) back to the server. 
Additionally, the FL system may employ either secure aggregation rules (e.g., Krum~\cite{yin2018byzantine} and FLDetector~\cite{zhang2022fldetector}) and/or fairness optimization strategies (e.g., \fedfb~\cite{zeng2021improving} and \fairfed~\cite{ezzeldin2021fairfed}) to defend against potential poisoning attacks and ensure model fairness.

On those malicious client devices, the adversary can launch the attack through three stages: (1) each client (e.g., the client $k$) follows the normal procedure by training the global model on their local datasets for a certain round of epochs, generating a benign model $\theta_{k,b}^{t}$; (2) the adversary uses the global model ($\theta_g^t$) to calculate the influence score of each local data sample from the privileged demographic group over the targeted group using Equation~\ref{eq:influ} and creates a local biasing dataset ($\mathcal{D}_{k}^{bias}$) that can reduce the model's classification accuracy for the targeted group; and (3) the adversary first identifies the redundant space (i.e., the neurons that remain relatively invariant during training) using LPR in the benign model update $\theta_{k,b}^{t}$, and then manipulates the neurons within this space using the created biasing dataset, leading to the poisoned model $\theta_{k,p}^{t}$. This poisoned model is sent to the server, where it is aggregated with updates from other clients to update the global model ($\theta_g^{t+1}$) for the next communication round. 
With this multi-step approach, the adversary can induce overfitting of the model to the privileged demographic group while decreasing accuracy for the targeted group, thereby exacerbating model bias, while the impact on its overall utility remains minimal.

\subsection{Design of \methodName}
\vspace{-1mm}
The crux of \methodName~lies in inducing bias without compromising the attack's persistence or stealthiness. Unlike conventional poisoning strategies that only attack the global model's utility, our attack employs an optimization-based approach that is more sophisticated and tailored for inducing bias in FL. The attack pipeline on each malicious client unfolds as:

\noindent(1) \textbf{Regular Local Training}: 
The adversary intends to introduce or exacerbate model bias by maintaining a high model utility for the privileged demographic group while decreasing the model accuracy for the targeted demographic group.
To ensure high utility for the privileged demographic, during the communication round $t$, each malicious client (e.g., client $k$) follows a similar procedure as benign clients to conduct training on their local dataset $\mathcal{D}_k$. The objective function can be represented as:
{\footnotesize\begin{equation}
\setlength{\abovedisplayskip}{3pt}
\setlength{\belowdisplayskip}{3pt}
\underset{\theta^t_{k,b}}{\min}\quad \frac{1}{|\mathcal{D}_k|} \sum_{i=1}^{|\mathcal{D}_k|} \mathcal{L}_c(f(x_i; \theta^t_{k,b}), y_i) \quad s.t. \phi(f,g_i) \leq \mu,
\end{equation}}
where $\phi(f,g_i)$ serves as a differentiable proxy for the fairness constraints (i.e., Demographic Parity used in \methodName), with $\mu$ representing a fairness tolerance parameter, and $|\mathcal{D}_k|$ is the number of local training samples.

\noindent(2) \textbf{Biasing Dataset Selection}:
The adversary then uses the current global model ($\theta_{g}^t$) to calculate the influence scores of local data samples from the privileged demographic group on the targeted demographic group, $\tau$, using Equation~\ref{eq:influ}. Upon calculating these influence scores, the adversary ranks the samples in ascending order and earmarks a predetermined fraction—denoted by $\kappa$ of the size of the privileged demographic group to create $\mathcal{D}_k^{bias}$. Choosing $\kappa$ is a strategic decision, that significantly impacts the attack's effectiveness. 


\noindent(3) \textbf{Model Poisoning}:
In FL, the adversary needs to ensure their model manipulations persist and remain effective amidst model updates aggregated from other clients on the server. While a straightforward strategy would be to introduce a significantly large local update, this method could jeopardize the attack's stealthiness and might be easily neutralized by secure aggregation rules. To tackle this, inspired by existing studies (e.g.,~\cite{li2019learn}) which show that certain neurons in a model are relatively redundant for model predictions and tend to remain invariant during training, \methodName embeds adversarial influences into this ``redundant'' space, aiming to establish a persistent and robust adversarial path within the model that can withstand the aggregation process in FL.

To effectively poison the model, the adversary initially identifies the model's redundant space in $\theta^t_{k,b}$. This is accomplished by utilizing Layer-Wise Relevance Propagation (described in Section~\ref{subsec:lrp}), which assigns relevance scores to each neuron, thereby highlighting their respective contributions to the model's primary classification task for the privileged group. Subsequently, the adversary proceeds to inject adversarial influences into this space by solving the following optimization problem with the selected biasing dataset $\mathcal{D}_k^{bias}$:
{\footnotesize\begin{equation}
\begin{split}
\setlength{\abovedisplayskip}{3pt}
\setlength{\belowdisplayskip}{3pt}
 \underset{\theta^t_{k,p}}{\min}  \frac{1}{|\mathcal{D}_k^{bias}|} \sum_{i=1}^{|\mathcal{D}_k^{bias}|} \mathcal{L}(l(x_i; \theta^t_{k,p}), y_i) &+ \gamma \sum_{\theta^*\in \theta^t_{k,p}} h(\theta^t_{k,b})(\Delta\theta^*)^2\\
     &+ \rho ||\theta^t_{k,p} - \theta^t_{k,b}||_2,
\end{split}
\vspace{-1mm}
\end{equation}}
where $\gamma$ and $\rho$ are weight factors for the two regularization terms, which are employed to ensure model poisoning occurs within the redundant space and to penalize the poisoned model update $(\theta^t_{k,p})$, which deviates much from the benign model update $(\theta^t_{k,b})$ produced in the initial step; $h(\theta^t_{k,b})$ represents the LRP score matrix of the model $\theta^t_{k,b}$, indicating the importance of neurons; and $\Delta\theta^*$ reflects the adjustments made during the model poisoning.

Solving this optimization problem enables the adversary to induce overfitting of the model to the privileged demographic group while decreasing accuracy for the targeted group, thus exacerbating model bias while maintaining overall model utility. The two regularization terms are designed to keep the poisonous updates closely aligned with benign updates and to confine modifications primarily to the redundant space, which tends to remain relatively stable during training. Such a strategy renders the attack more covert and helps it endure the aggregation process on the server in FL.

\vspace{-3mm}
\section{Evaluation}
\vspace{-1mm}
\subsection{Federated Datasets}
\vspace{-1mm}
We evaluate the proposed \methodName~using the following three datasets in non-IID settings:


\noindent(1) \textit{CelebA}~\cite{liu2018large}: A collection of $200k$ celebrity face images from the Internet that have been manually annotated. The dataset has up to 40 labels, each of which is binary-valued. For CelebA, each subject's gender (male or female) is a sensitive attribute. (2) \textit{Adult Income}~\cite{Dheeru2017}: A tabular dataset that is widely investigated in machine learning fairness literature. It contains $48,842$ samples with $14$ attributes. In this dataset, race (white or non-white) is used as the sensitive attribute. (3) \textit{UTK Faces}~\cite{zhifei2017cvpr}: A large-scale face dataset with more than 20,000 face images with annotations of age, gender, and ethnicity. Race (white or non-white) is used as the sensitive attribute in this dataset.

To show the real-world implications, we also apply \methodName to the \textit{MovieLens 1M} dataset~\cite{harper2015movielens} (a movie recommendation system).

\subsection{Evaluation Metrics}
\vspace{-1mm}

\noindent(1) \textit{Equal Opportunity Difference (EOD)}:
We use the EOD of each sensitive group to measure group fairness. Specifically, $\textup{EOD} = |Pr\{\hat{Y}=1|G=0, Y=1\} - Pr\{\hat{Y}=1|G=1, Y=1\}|$.

\noindent(2) \textit{Demographic Parity Difference (DPD)}: DPD is another metric used for measuring group fairness, which is calculated as $\textup{DPD} = \left|Pr\{\hat{Y}=1|G=a\}-Pr\{\hat{Y}=1\}\right|$.

\noindent(3) \textit{Utility}: In our experiments, we use the model's prediction accuracy
to quantify global model utility. 

\begin{table*}[t]
\begin{center}
\resizebox{0.87\linewidth}{!}{
\begin{tabular}{@{}ccccccccccc@{}}
\toprule
         \multirow{4}{6em}{\centering\textbf{Fairness Optimization}} &\multirow{4}{3em}{\textbf{Attack}} &\multicolumn{3}{c}{\textbf{CelebA} ($\epsilon=0.1$)} &\multicolumn{3}{c}{\textbf{Adult Income} ($\epsilon=0.2$)} &\multicolumn{3}{c}{\textbf{UTK Faces} ($\epsilon=0.2$)}\\ \cmidrule(r){3-5} \cmidrule(r){6-8} \cmidrule(r){9-11}
         
        &&{\textbf{EOD} ($\downarrow$)} &{\textbf{DPD}($\downarrow$)} &\multirow{2}{2.5em}{\textbf{Utility}} &{\textbf{EOD} ($\downarrow$)} &{\textbf{DPD}($\downarrow$)} &\multirow{2}{2.5em} {\textbf{Utility}} &{\textbf{EOD} ($\downarrow$)} &{\textbf{DPD}($\downarrow$)} &\multirow{2}{2.5em}{\textbf{Utility}} \\ \cmidrule(r){3-4}   \cmidrule(r){6-7} \cmidrule(r){9-10}     
    
        & &\textbf{Gender} &\textbf{Gender} &\Gape[1ex][1ex]{($\uparrow$)} &\textbf{Race} &\textbf{Race} & \Gape[1ex][1ex]{($\uparrow$)} &\textbf{Race} & \textbf{Race} &\Gape[1ex][1ex]{($\uparrow$)}\\\midrule

\multirow{2}{4em}{\fedavg} 
                    &No Attack      &0.23 &0.21 &\textbf{91}\% &0.25 &0.27 &\textbf{83}\% &0.24 &0.22 &\textbf{87}\%\\
                    &Gradient-based &0.25 &0.23 &85\% &0.27 &0.27 &79\% &0.29 &0.31 &82\%\\
 ~\cite{mcmahan2017communication} &Anchoring-based      &0.24 &0.22 &81\% &0.27 &0.29 &77\% &0.27 &0.28 &81\%\\
                    &\textbf{~\methodName}   &\textbf{0.41} &\textbf{0.43} &88\% &\textbf{0.41} &\textbf{0.44} &80\% &\textbf{0.38} &\textbf{0.34} &84\%\\\midrule
     
 \multirow{3}{3em}{\qffl} 
                   &No Attack      &0.19 &0.20 &\textbf{89}\% &0.22 &0.24 &\textbf{82}\% &0.18 &0.21 &\textbf{84}\%\\
                   &Gradient-based &0.21 &0.21 &84\% &0.26 &0.28 &79\% &0.22 &0.25 &80\%\\
~\cite{li2019fair} &Anchoring-based      &0.21 &0.23 &83\% &0.24 &0.24 &76\% &0.21 &0.21 &79\%\\
                   &\textbf{~\methodName}   &\textbf{0.36} &\textbf{0.39} &85\% &\textbf{0.39} &\textbf{0.38} &79\% &\textbf{0.33} &\textbf{0.30} & 81\%\\\midrule
     
 \multirow{2}{5.5em}{\gifair} 
                      &No Attack      &0.19 &0.19 &\textbf{88}\% &0.21 &0.23 &\textbf{82}\% &0.17 &0.20 &\textbf{83}\%\\
                      &Gradient-based &0.21 &0.20 &84\% &0.25 &0.28 &78\% &0.20 &0.21 &76\%\\
~\cite{yue2021gifair} &Anchoring-based      &0.21 &0.22 &81\% &0.22 &0.24 &76\% &0.20 &0.19 &73\%\\
                      &\textbf{~\methodName}   &\textbf{0.33} &\textbf{0.37} &84\% &\textbf{0.38} &\textbf{0.36} &79\% &\textbf{0.31} &\textbf{0.28} &78\% \\\midrule
     
  \multirow{3}{4em}{\fairfed} 
                            &No Attack      &0.16 &0.16 &\textbf{87}\% &0.18 &0.19 &\textbf{80}\% &0.17 &0.15 &\textbf{82}\%\\
                            &Gradient-based &0.19 &0.20 &83\% &0.23 &0.25 &75\% &0.23 &0.21 &72\%\\
~\cite{ezzeldin2021fairfed} &Anchoring-based      &0.21 &0.19 &79\% &0.19 &0.21 &72\% &0.20 &0.19 &70\%\\
                            &\textbf{~\methodName}   &\textbf{0.31} &\textbf{0.26} &84\% &\textbf{0.29} &\textbf{0.30} &78\% &\textbf{0.28} &\textbf{0.28} &76\% \\\midrule
     
 \multirow{2}{3.5em}{\fedfb} 
                          &No Attack      &0.15 &0.16 &\textbf{84}\% &0.19 &0.19 &\textbf{79}\% &0.16 &0.17 &\textbf{82}\%\\
                          &Gradient-based &0.21 &0.23 &79\% &0.25 &0.25 &72\% &0.20 &0.22 &75\%\\
~\cite{zeng2021improving}&Anchoring-based       &0.19 &0.18 &76\% &0.24 &0.25 &70\% &0.19 &0.20 &72\%\\
                         &\textbf{~\methodName}    &\textbf{0.31} &\textbf{0.29} &81\% &\textbf{0.32} &\textbf{0.33} &75\% &\textbf{0.30} &\textbf{0.30} &77\%\\ \bottomrule
\end{tabular}
}
\end{center}
\vspace{-2mm}
\caption{Group fairness comparison of different fairness optimization methods under different attack scenarios.} \label{table-1}
\vspace{-2mm}
\end{table*}

\subsection{Fairness Attack Baselines}
\vspace{-1mm}
Since no fairness attack specifically designed for FL currently exists, we adapted two fairness attacks originally designed for centralized learning (i.e., gradient-based~\cite{solans2021poisoning} and anchoring-based attack~\cite{mehrabi2021exacerbating}) to FL settings to demonstrate the superiority of \methodName. Specifically, to introduce or exacerbate model bias, the gradient-based attack employs a bi-level optimization process to inject a small fraction of poisoning points into the training data, while the anchoring-based attack strategically introduces poisoned data points near the targeted demographic group, sharing the same demographic characteristics but with opposite labels.
In FL settings, we adapted these attacks by enabling malicious clients to employ them locally during their training processes. 
Unlike centralized learning, since we cannot measure the fairness impact of each local data sample on the global model, these attacks adapted for FL primarily target degrading the fairness level within their respective local datasets, rather than the global model as a whole.


\subsection{Fairness Optimization Strategies}
\vspace{-1mm}
To evaluate the effectiveness of our attack under certain fairness optimization 
strategies applied in FL, besides \fedavg~\cite{mcmahan2017communication}, we 
also, adopt the following state-of-the-art FL fairness optimization methods to 
evaluate our attack's effectiveness: \textit{(i)} \qffl~\cite{li2019fair}: \qffl 
is one of the client-fairness-based methods, aiming to equalize the accuracies 
of all the clients by dynamically reweighting the aggregation, favoring the 
clients with poor performance. \textit{(ii)} \gifair~\cite{yue2021gifair}: 
\gifair aims to achieve client fairness using regularization terms to penalize 
the spread in the loss. \textit{(iii)} \fedfb~\cite{zeng2021improving}: \fedfb 
is a group fairness method, where they have fitted the concept of the fair batch 
from centralized learning into FL by leveraging the shared group-specific 
positive prediction rate for each client. \textit{(iv)} 
\fairfed~\cite{ezzeldin2021fairfed}: \fairfed is a group fairness method that 
can improve both local and group fairness. It employs \fedavg~and a fairness-
based re-weighting mechanism to account for the mismatch between global fairness 
measure and local fairness measure.

\subsection{Attack Performance}
\vspace{-1mm}
\textbf{Attack Performance under \fedavg.}
As shown in Table~\ref{table-1}, if no attack is present under \fedavg, the learned global model tends to be biased against certain demographic groups already (e.g., EOD and DPD on the CelebA dataset are 0.23 and 0.21, respectively), while the model utility is at a relatively good level (e.g., $91\%$ on the CelebA dataset). Introducing an attack amplifies this inherent unfairness, as demonstrated in our evaluations where each participating client has an $\epsilon$ probability of being malicious. 
The results indicate that compared to the two fairness attacks, \methodName is capable of introducing significantly greater model bias while maintaining a minimal impact on its utility.
In the \fedavg setting, without employing any fairness optimizations, these baseline attacks only slightly increase model bias.
This is likely due to the adversarial updates being overshadowed by benign updates, leading to a catastrophic forgetting of the adversarial update (for instance, EOD values for the gradient-based and anchoring-based attacks on the CelebA dataset are just 0.25 and 0.24, respectively). However,~\methodName significantly impacts the global model fairness (e.g., EOD and DPD values on the CelebA dataset are 0.41 and 0.43, respectively), meanwhile keeping its utility high.  

\noindent\textbf{Attack Performance under Fairness Optimization.} 
To validate whether our attack can remain effective under existing fair optimization strategies in FL, we evaluate our attack under four state-of-the-art fair FL methods, including \qffl~\cite{li2019fair}, \gifair~\cite{yue2021gifair}, \fairfed~\cite{ezzeldin2021fairfed} and \fedfb~\cite{zeng2021improving}. The results, as shown in Table~\ref{table-1}, indicate that while baseline attacks such as gradient-based and anchoring-based attacks yield a slight degradation in performance, \methodName significantly undermines fairness, even in the face of these advanced optimization strategies.
For instance, when applying \qffl to the UTK Faces dataset, our attack results in an EOD of $0.33$, signifying that the privileged racial group (white) is $33\%$ more likely to receive correct classifications. Similarly, a DPD of $0.30$ indicates a $30\%$ higher likelihood for the privileged group to be positively classified. We also observe that~\methodName is relatively more effective against \qffl and \gifair than against \fairfed and \fedfb. This is likely because \fairfed and \fedfb implement fairness constraints at the local model level, enhancing each local model's fairness. In contrast, \qffl and \gifair apply fairness during server-side aggregation, making them more susceptible to exploitation by~\methodName.
\begin{figure}[t]
\begin{minipage}[t]{0.5\linewidth}
    \centering
    \includegraphics[width=\linewidth]{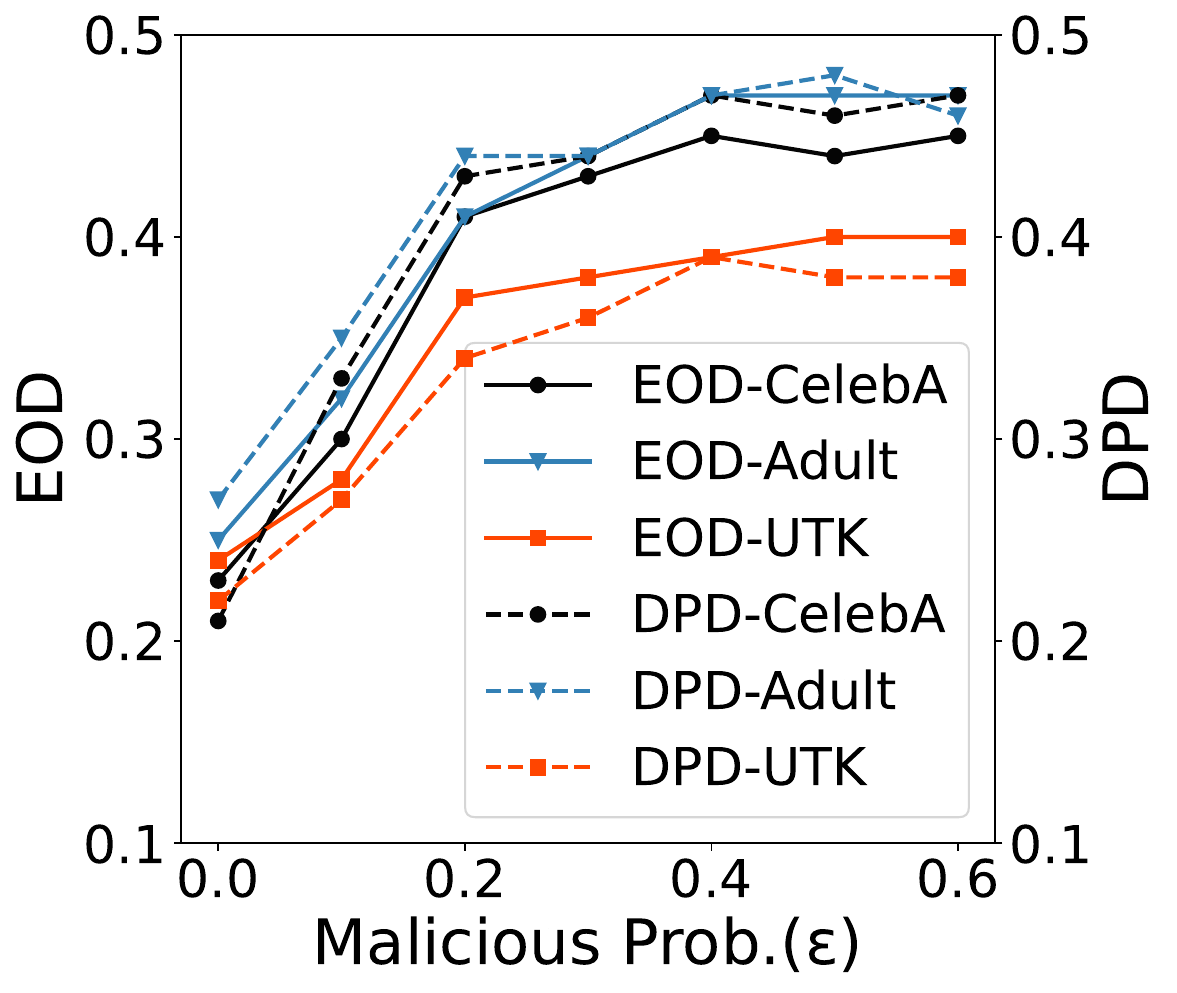}
    \vspace{-3mm}
    \caption{Impact of malicious participant probability.}
    \vspace{-2mm}
    \label{fig:percentage}
\end{minipage}
\hspace{0.01cm}
\begin{minipage}[t]{0.47\linewidth} 
    \centering
    \includegraphics[width=\linewidth]{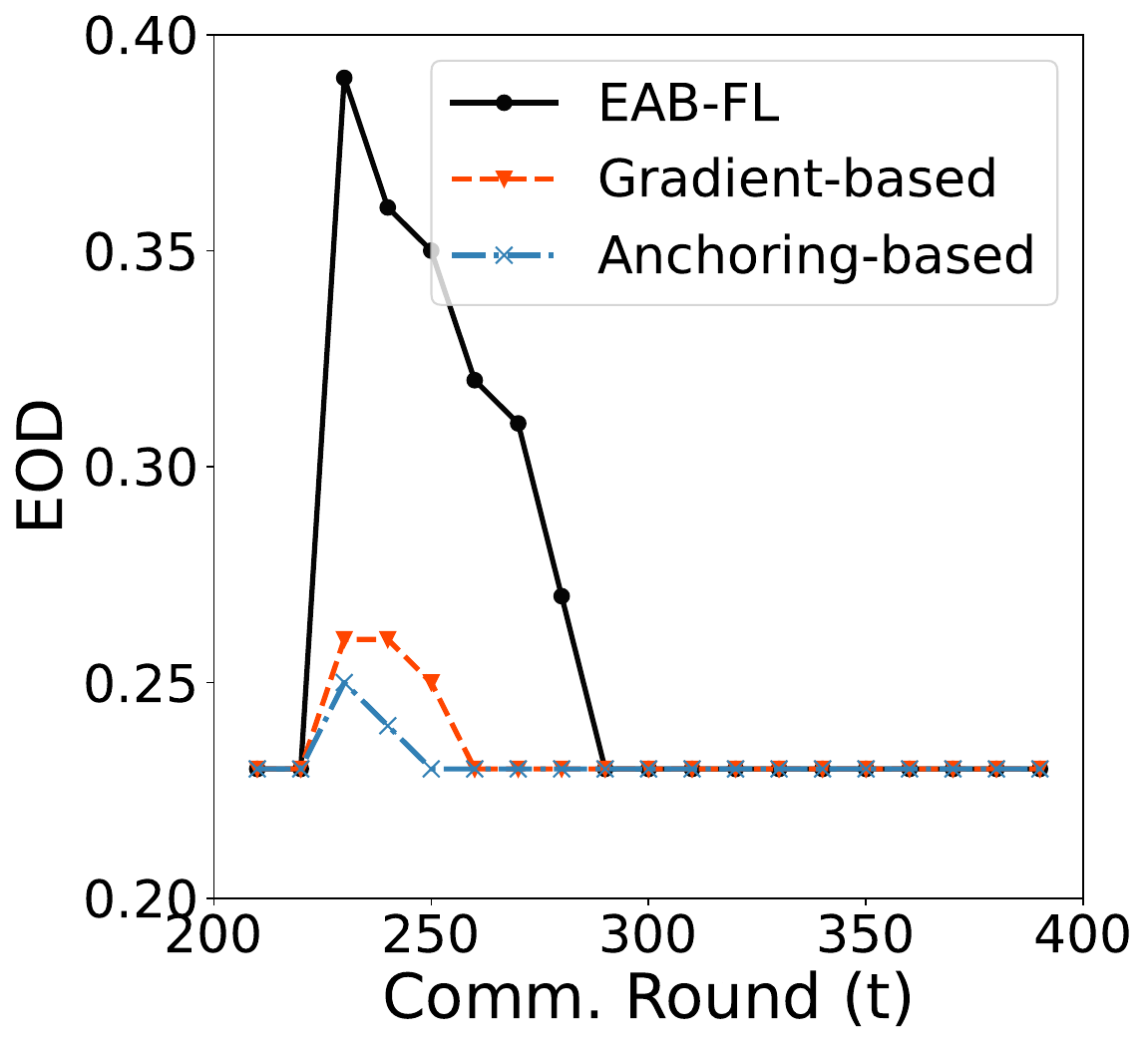}
    \vspace{-3mm}
    \caption{Attack persistence evaluation.}
    \vspace{-2mm}
    \label{fig:persistance}
\end{minipage}   
\end{figure}

\begin{figure}[b]
    \centering
    \vspace{-2mm}
    \begin{minipage}[t]{0.49\linewidth}
        \centering
        \includegraphics[width=\linewidth]{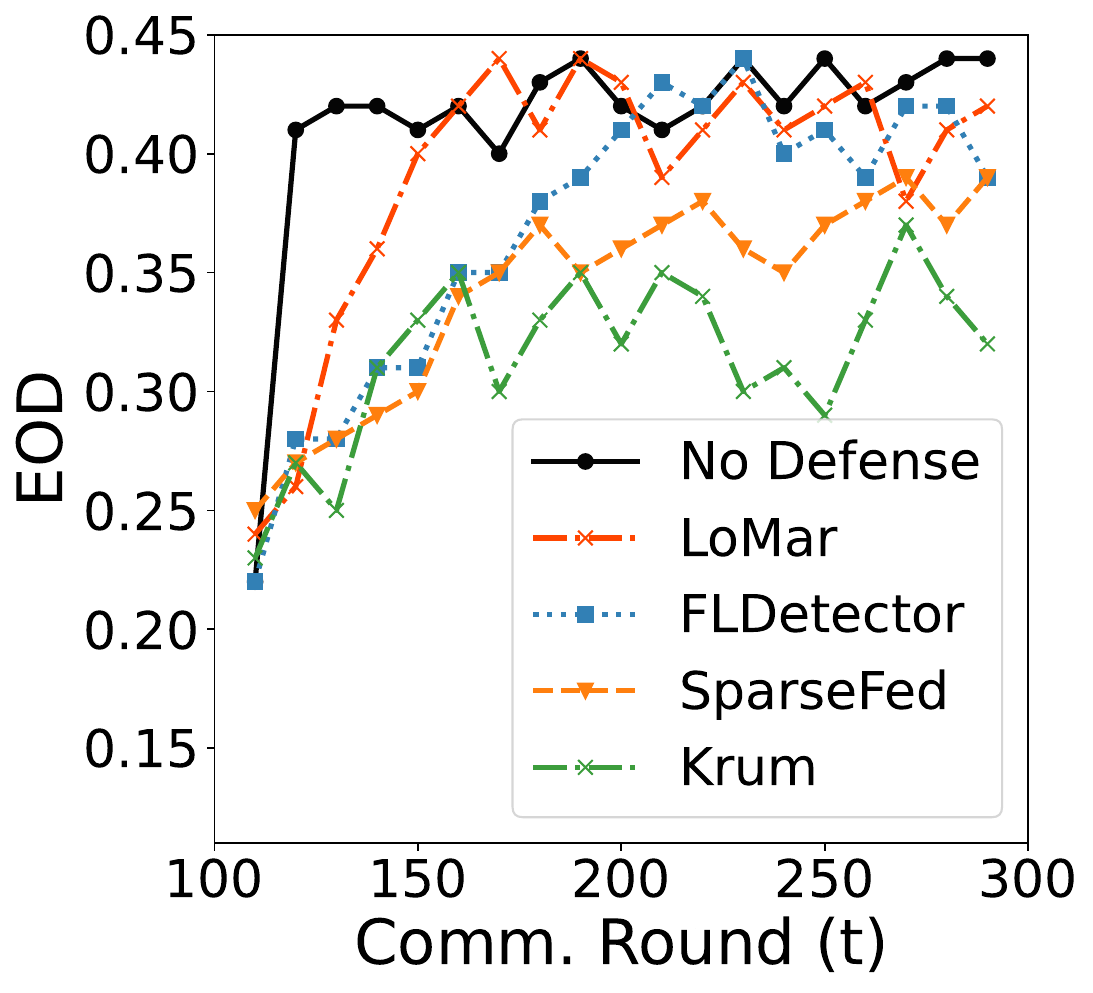}
        \vspace{-2mm}
        \subcaption{}
    \end{minipage}
    \hspace{0.01cm}
    \begin{minipage}[t]{0.49\linewidth} 
        \centering
        \includegraphics[width=\linewidth]{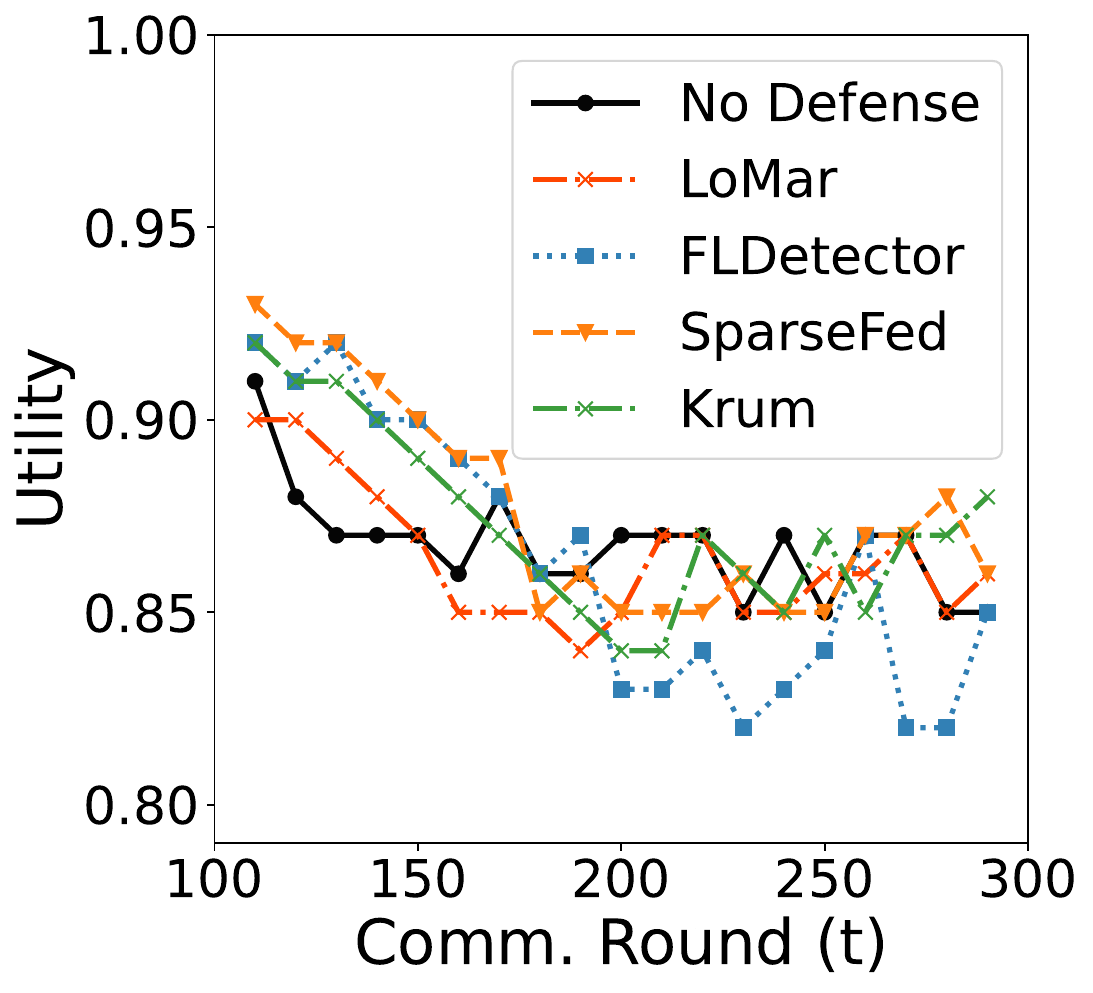}
        \vspace{-2mm}
        \subcaption{}
    \end{minipage}
    \caption{Attack effectiveness against secure aggregations.}
	\label{fig:stealth}
    \label{fig:combined}
\end{figure}

\noindent\textbf{Impact of Malicious Participant Availability ($\epsilon$).} 
As shown in Figure~\ref{fig:percentage}, we can see that the fairness measures (i.e., EOD and DPD) significantly increase from $0.23$ and $0.21$ to $0.41$ and $0.43$ for the CelebA data with the increasing probability of participating clients being malicious ($\epsilon$, from $0$ to $0.3$). However, further increasing $\epsilon$ does not have a proportional impact on the model's fairness. This is due to the constraints on the model weights that make the malicious updates appear less suspicious to the server and prevent the client's local model from being trivial. We can observe a similar trend for other datasets as well, like for the Adult Income data dataset, the saturation comes at $0.4$, and for UTK faces it comes at $0.3$.

\noindent\textbf{Attack Persistence.}
Attack persistence allows us to measure the durability of the attack's impact after the removal of malicious clients from the training process. To evaluate this, we conduct a single-shot attack scenario on the CelebA dataset, where the attack is initiated only once at round $t = 230$. For this round, we assume that all participating clients are malicious. Figure~\ref{fig:persistance} presents a comparative analysis of \methodName and other baseline fairness attacks under this scenario, which indicates that~\methodName maintains a high level of attack success (high EOD), over an extended period of model aggregation. The key to the proposed attack's sustained effectiveness lies in its strategic placement of poisoning neurons within the neural network's redundant space. By embedding the adversarial influence in these less dynamic areas of the network, we significantly reduce the likelihood of these neurons being altered during the training of the main task, thus ensuring the longevity of the attack's impact.

\noindent\textbf{Effectiveness against Secure Aggregations.}
We evaluate the effectiveness of \methodName~against four robust aggregation rules: SparseFed~\cite{panda2022sparsefed}, Krum~\cite{yin2018byzantine}, LoMar~\cite{li2021lomar}, and FLDetector~\cite{zhang2022fldetector} on the CelebA dataset.
As shown in Figure~\ref{fig:stealth}(a), our proposed~\methodName can bypass the norm-bounded defense method (SparseFed) and achieves significantly high unfairness (e.g., EOD is over $0.35$) in the global model. Although Krum, effective at detecting aberrant models, sometimes rejects our attack's poisonous updates, the adversary still maintains a much higher EOD (e.g., over $0.3$ for most cases) compared to scenarios without attacks. Model-similarity-based methods, such as LoMar and FLDetector, fail to detect \methodName, as the poisonous updates closely resemble benign updates, constrained by the L2 norm.
Figure~\ref{fig:stealth}(b) shows that \methodName retains high model utility, irrespective of the use of secure aggregations.
\begin{table}[t]

\begin{center}
\resizebox{0.98\linewidth}{!}{
\begin{tabular}{@{}cccc@{}}
\toprule
         \multirow{2}{4em}{\textbf{Attack}}  &{\textbf{EOD} ($\downarrow$)} &{\textbf{DPD}($\downarrow$)} &\multirow{2}{3.5em}{\textbf{Utility}}\\ \cmidrule(r){2-3}
         &\textbf{Gender} &\textbf{Gender} &\Gape[1ex][1ex]{($\uparrow$)}\\          \midrule
    DeSMP~\cite{hossain2021desmp} &0.24 &0.21 &52\%     \\
    MPAF~\cite{cao2022mpaf} &0.28 &0.30 &49\%\\
    Sign Flipping~\cite{li2019rsa} &0.25 &0.23 &55\%\\
    DYN-OPT~\cite{shejwalkar2021manipulating} &0.29 &0.31 &53\%\\
    BDB~\cite{shejwalkar2022back} &0.32 &0.29 &51\%\\
    \textbf{\methodName} &\textbf{0.45} &\textbf{0.50} &\textbf{83\%}\\ \bottomrule
\end{tabular}
}
\vspace{-2mm}
\end{center}
\caption {Comparison of~\methodName with other poisoning attacks.} 
\label{table-2} 
\vspace{-2mm}
\end{table}
\noindent\textbf{Comparison with Other Poisoning Attacks.}
To show the attack's unique impact on the model's fairness and utility, we compare \methodName~with various state-of-the-art poisoning attacks (i.e., DeSMP~\cite{hossain2021desmp}, MPAF~\cite{cao2022mpaf}, Sign Flipping~\cite{li2019rsa}, DYN-OPT~\cite{shejwalkar2021manipulating}, and BDB~\cite{shejwalkar2022back}) in FL on the CelebA dataset. As shown in Table~\ref{table-2}, we observe that our attack achieves the highest overall impact on the model's fairness (EDO and DPD are as high as 0.45 and 0.50, respectively), and the model's utility is still at a relatively high level (i.e., over $83\%$). This indicates our attack's effectiveness in exacerbating group unfairness while preserving the utility of the global model. Conversely, the other poisoning attacks all have a negligible impact on fairness, i.e., less than $0.32$ and $0.31$ in EOD and DPD, while degrading utility to as low as $49\%$, mainly due to the non-uniform changes in local data or model.

\begin{table}[t]

\begin{center}
\resizebox{0.98\linewidth}{!}{
\begin{tabular}{@{}ccccc@{}}
\toprule
        \textbf{Attack}  &{\textbf{No Attack}} &{\textbf{Gradient-based}} &{\textbf{Anchoring-based}} &{\textbf{\methodName}}\\       
         \midrule
         \textbf{Time (s)} &1.92 &19.13 &4.7 &\textbf{3.36}\\
\bottomrule
\end{tabular}
}
\vspace{-2mm}
\end{center}
\caption {Time Consumption Analysis.} 
\label{table-3} 
\vspace{-2mm}
\end{table}
\noindent\textbf{Time Complexity Analysis.} 
Table~\ref{table-3} shows the average time required to successfully attack the global model per communication round on the CelebA dataset using an Nvidia Quadro A100 GPU. The results show that \methodName needs a slightly higher computation time compared to the no-attack (\fedavg) scenario.
To demonstrate \methodName's feasibility on lower computational power devices like smartphones and laptops, we estimated the required time by comparing their GPUs' FLOPS.
The A100 GPU, chips used in recent Apple smartphones (e.g., Apple A17 Bionic), and commonly used chips in laptops (e.g., Intel Core i7 13700) have FLOPS ratings of $9.7$ TFLOPS~\cite{nvidiaNVIDIAA100}, $2.15$ TFLOPS~\cite{cpumonkeyAppleCores}, and $0.82$ TFLOPS~\cite{cpumonkeyIntelCore}, respectively. The estimated consumed time of~\methodName on smartphones is $15.16$ seconds and $39.74$ seconds on laptops, which confirms the feasibility of launching \methodName on these edge devices.

\vspace{-3mm}
\section{Conclusion}
\vspace{-1mm}
In this work, we propose a new type of model poisoning attack,~\methodName, in FL settings, with a focus on exacerbating group unfairness while maintaining a good level of model utility. 
The effectiveness and efficiency of the proposed attack are demonstrated through extensive experiments on three datasets in various FL settings. The results of this study highlight the importance of fully understanding the attack surfaces of current FL systems and the need for corresponding mitigations to improve their resilience against such attacks. 

\bibliographystyle{named}
\bibliography{ijcai24}

\begin{thebibliography}{}

\bibitem[\protect\citeauthoryear{Abay \bgroup \em et al.\egroup }{2020}]{abay2020mitigating}
Annie Abay, Yi~Zhou, Nathalie Baracaldo, Shashank Rajamoni, Ebube Chuba, and Heiko Ludwig.
\newblock Mitigating bias in federated learning.
\newblock {\em arXiv preprint arXiv:2012.02447}, 2020.

\bibitem[\protect\citeauthoryear{Bach \bgroup \em et al.\egroup }{2015}]{bach2015pixel}
Sebastian Bach, Alexander Binder, Gr{\'e}goire Montavon, Frederick Klauschen, Klaus-Robert M{\"u}ller, and Wojciech Samek.
\newblock On pixel-wise explanations for non-linear classifier decisions by layer-wise relevance propagation.
\newblock {\em PloS one}, 10(7):e0130140, 2015.

\bibitem[\protect\citeauthoryear{Bagdasaryan \bgroup \em et al.\egroup }{2020}]{bagdasaryan2020backdoor}
Eugene Bagdasaryan, Andreas Veit, Yiqing Hua, Deborah Estrin, and Vitaly Shmatikov.
\newblock How to backdoor federated learning.
\newblock In {\em International conference on artificial intelligence and statistics}, pages 2938--2948. PMLR, 2020.

\bibitem[\protect\citeauthoryear{Bhagoji \bgroup \em et al.\egroup }{2019}]{bhagoji2019analyzing}
Arjun~Nitin Bhagoji, Supriyo Chakraborty, Prateek Mittal, and Seraphin Calo.
\newblock Analyzing federated learning through an adversarial lens.
\newblock In {\em International Conference on Machine Learning}, pages 634--643. PMLR, 2019.

\bibitem[\protect\citeauthoryear{Biddle}{2017}]{biddle2017adverse}
Dan Biddle.
\newblock {\em Adverse impact and test validation: A practitioner's guide to valid and defensible employment testing}.
\newblock Routledge, 2017.

\bibitem[\protect\citeauthoryear{Cao and Gong}{2022}]{cao2022mpaf}
Xiaoyu Cao and Neil~Zhenqiang Gong.
\newblock Mpaf: Model poisoning attacks to federated learning based on fake clients.
\newblock In {\em Proceedings of the IEEE/CVF Conference on Computer Vision and Pattern Recognition}, pages 3396--3404, 2022.

\bibitem[\protect\citeauthoryear{Chhabra \bgroup \em et al.\egroup }{2023}]{chhabra2022robust}
Anshuman Chhabra, Peizhao Li, Prasant Mohapatra, and Hongfu Liu.
\newblock Robust fair clustering: A novel fairness attack and defense framework.
\newblock {\em International Conference on Learning Representations (ICLR)}, 2023.

\bibitem[\protect\citeauthoryear{Chu \bgroup \em et al.\egroup }{2021}]{chu2021fedfair}
Lingyang Chu, Lanjun Wang, Yanjie Dong, Jian Pei, Zirui Zhou, and Yong Zhang.
\newblock Fedfair: Training fair models in cross-silo federated learning.
\newblock {\em arXiv preprint arXiv:2109.05662}, 2021.

\bibitem[\protect\citeauthoryear{Clauset \bgroup \em et al.\egroup }{2009}]{clauset2009power}
Aaron Clauset, Cosma~Rohilla Shalizi, and Mark~EJ Newman.
\newblock Power-law distributions in empirical data.
\newblock {\em SIAM review}, 51(4):661--703, 2009.

\bibitem[\protect\citeauthoryear{Cpu-Monkey}{2024a}]{cpumonkeyAppleCores}
Cpu-Monkey.
\newblock {A}pple {A}17 {P}ro (6 {G}{P}{U} {C}ores) {B}enchmark, {T}est and specs --- cpu-monkey.com.
\newblock \url{https://www.cpu-monkey.com/en/igpu-apple_a17_pro_6_gpu_cores}, 2024.
\newblock [Accessed 17-01-2024].

\bibitem[\protect\citeauthoryear{Cpu-Monkey}{2024b}]{cpumonkeyIntelCore}
Cpu-Monkey.
\newblock {I}ntel {C}ore i7-13700 {B}enchmark, {T}est and specs --- cpu-monkey.com.
\newblock \url{https://www.cpu-monkey.com/en/cpu-intel_core_i7_13700}, 2024.
\newblock [Accessed 17-01-2024].

\bibitem[\protect\citeauthoryear{Du \bgroup \em et al.\egroup }{2021}]{du2021fairness}
Wei Du, Depeng Xu, Xintao Wu, and Hanghang Tong.
\newblock Fairness-aware agnostic federated learning.
\newblock In {\em Proceedings of the 2021 SIAM International Conference on Data Mining (SDM)}, pages 181--189. SIAM, 2021.

\bibitem[\protect\citeauthoryear{Dua and Graff}{2017}]{Dheeru2017}
Dheeru Dua and Casey Graff.
\newblock Uci machine learning repository.
\newblock \url{https://archive.ics.uci.edu/ml/datasets/adult}, 2017.
\newblock [Online; accessed 17-December-2021].

\bibitem[\protect\citeauthoryear{Dwork \bgroup \em et al.\egroup }{2012}]{dwork2012fairness}
Cynthia Dwork, Moritz Hardt, Toniann Pitassi, Omer Reingold, and Richard Zemel.
\newblock Fairness through awareness.
\newblock In {\em Proceedings of the 3rd innovations in theoretical computer science conference}, pages 214--226, 2012.

\bibitem[\protect\citeauthoryear{Ezzeldin \bgroup \em et al.\egroup }{2023}]{ezzeldin2021fairfed}
Yahya~H Ezzeldin, Shen Yan, Chaoyang He, Emilio Ferrara, and A~Salman Avestimehr.
\newblock Fairfed: Enabling group fairness in federated learning.
\newblock {\em Proceedings of the AAAI Conference on Artificial Intelligence}, 37(6):7494--7502, 2023.

\bibitem[\protect\citeauthoryear{Hardt \bgroup \em et al.\egroup }{2016}]{hardt2016equality}
Moritz Hardt, Eric Price, and Nati Srebro.
\newblock Equality of opportunity in supervised learning.
\newblock {\em Advances in neural information processing systems}, 29, 2016.

\bibitem[\protect\citeauthoryear{Harper and Konstan}{2015}]{harper2015movielens}
F~Maxwell Harper and Joseph~A Konstan.
\newblock The movielens datasets: History and context.
\newblock {\em Acm transactions on interactive intelligent systems (tiis)}, 5(4):1--19, 2015.

\bibitem[\protect\citeauthoryear{He \bgroup \em et al.\egroup }{2016}]{he2016deep}
Kaiming He, Xiangyu Zhang, Shaoqing Ren, and Jian Sun.
\newblock Deep residual learning for image recognition.
\newblock In {\em Proceedings of the IEEE conference on computer vision and pattern recognition}, pages 770--778, 2016.

\bibitem[\protect\citeauthoryear{He \bgroup \em et al.\egroup }{2017}]{he2017neural}
Xiangnan He, Lizi Liao, Hanwang Zhang, Liqiang Nie, Xia Hu, and Tat-Seng Chua.
\newblock Neural collaborative filtering.
\newblock In {\em Proceedings of the 26th international conference on world wide web}, pages 173--182, 2017.

\bibitem[\protect\citeauthoryear{Hossain \bgroup \em et al.\egroup }{2021}]{hossain2021desmp}
Md~Tamjid Hossain, Shafkat Islam, Shahriar Badsha, and Haoting Shen.
\newblock Desmp: Differential privacy-exploited stealthy model poisoning attacks in federated learning.
\newblock In {\em 2021 17th International Conference on Mobility, Sensing and Networking (MSN)}, pages 167--174. IEEE, 2021.

\bibitem[\protect\citeauthoryear{Jacot \bgroup \em et al.\egroup }{2018}]{jacot2018neural}
Arthur Jacot, Franck Gabriel, and Cl{\'e}ment Hongler.
\newblock Neural tangent kernel: Convergence and generalization in neural networks.
\newblock {\em Advances in neural information processing systems}, 31, 2018.

\bibitem[\protect\citeauthoryear{Konečný \bgroup \em et al.\egroup }{2016}]{konevcny2016federated}
Jakub Konečný, H.~Brendan McMahan, Felix~X. Yu, Peter Richtarik, Ananda~Theertha Suresh, and Dave Bacon.
\newblock Federated learning: Strategies for improving communication efficiency.
\newblock In {\em NIPS Workshop on Private Multi-Party Machine Learning}, 2016.

\bibitem[\protect\citeauthoryear{Li \bgroup \em et al.\egroup }{2019a}]{li2019rsa}
Liping Li, Wei Xu, Tianyi Chen, Georgios~B Giannakis, and Qing Ling.
\newblock Rsa: Byzantine-robust stochastic aggregation methods for distributed learning from heterogeneous datasets.
\newblock In {\em Proceedings of the AAAI Conference on Artificial Intelligence}, volume~33, pages 1544--1551, 2019.

\bibitem[\protect\citeauthoryear{Li \bgroup \em et al.\egroup }{2019b}]{li2019learn}
Xilai Li, Yingbo Zhou, Tianfu Wu, Richard Socher, and Caiming Xiong.
\newblock Learn to grow: A continual structure learning framework for overcoming catastrophic forgetting.
\newblock In {\em International Conference on Machine Learning}, pages 3925--3934. PMLR, 2019.

\bibitem[\protect\citeauthoryear{Li \bgroup \em et al.\egroup }{2020}]{li2019fair}
Tian Li, Maziar Sanjabi, Ahmad Beirami, and Virginia Smith.
\newblock Fair resource allocation in federated learning.
\newblock {\em International Conference on Learning Representations (ICLR)}, 2020.

\bibitem[\protect\citeauthoryear{Li \bgroup \em et al.\egroup }{2021a}]{li2021ditto}
Tian Li, Shengyuan Hu, Ahmad Beirami, and Virginia Smith.
\newblock Ditto: Fair and robust federated learning through personalization.
\newblock In {\em International Conference on Machine Learning}, pages 6357--6368. PMLR, 2021.

\bibitem[\protect\citeauthoryear{Li \bgroup \em et al.\egroup }{2021b}]{li2021lomar}
Xingyu Li, Zhe Qu, Shangqing Zhao, Bo~Tang, Zhuo Lu, and Yao Liu.
\newblock Lomar: A local defense against poisoning attack on federated learning.
\newblock {\em IEEE Transactions on Dependable and Secure Computing}, 2021.

\bibitem[\protect\citeauthoryear{Li \bgroup \em et al.\egroup }{2022}]{li2022learning}
Henger Li, Xiaolin Sun, and Zizhan Zheng.
\newblock Learning to attack federated learning: A model-based reinforcement learning attack framework.
\newblock {\em Advances in Neural Information Processing Systems}, 35:35007--35020, 2022.

\bibitem[\protect\citeauthoryear{Liu \bgroup \em et al.\egroup }{2018}]{liu2018large}
Ziwei Liu, Ping Luo, Xiaogang Wang, and Xiaoou Tang.
\newblock Large-scale celebfaces attributes (celeba) dataset.
\newblock {\em Retrieved August}, 15(2018):11, 2018.

\bibitem[\protect\citeauthoryear{Liu \bgroup \em et al.\egroup }{2023}]{liu2023efficient}
Tao Liu, Zhi Wang, Hui He, Wei Shi, Liangliang Lin, Ran An, and Chenhao Li.
\newblock Efficient and secure federated learning for financial applications.
\newblock {\em Applied Sciences}, 13(10):5877, 2023.

\bibitem[\protect\citeauthoryear{Lyu \bgroup \em et al.\egroup }{2020}]{lyu2020collaborative}
Lingjuan Lyu, Xinyi Xu, Qian Wang, and Han Yu.
\newblock Collaborative fairness in federated learning.
\newblock In {\em Federated Learning}, pages 189--204. Springer, 2020.

\bibitem[\protect\citeauthoryear{McMahan \bgroup \em et al.\egroup }{2017}]{mcmahan2017communication}
Brendan McMahan, Eider Moore, Daniel Ramage, Seth Hampson, and Blaise~Aguera y~Arcas.
\newblock Communication-efficient learning of deep networks from decentralized data.
\newblock In {\em Artificial intelligence and statistics}, pages 1273--1282. PMLR, 2017.

\bibitem[\protect\citeauthoryear{Mehrabi \bgroup \em et al.\egroup }{2021}]{mehrabi2021exacerbating}
Ninareh Mehrabi, Muhammad Naveed, Fred Morstatter, and Aram Galstyan.
\newblock Exacerbating algorithmic bias through fairness attacks.
\newblock In {\em Proceedings of the AAAI Conference on Artificial Intelligence}, volume~35, pages 8930--8938, 2021.

\bibitem[\protect\citeauthoryear{Mohri \bgroup \em et al.\egroup }{2019}]{mohri2019agnostic}
Mehryar Mohri, Gary Sivek, and Ananda~Theertha Suresh.
\newblock Agnostic federated learning.
\newblock In {\em International Conference on Machine Learning}, pages 4615--4625. PMLR, 2019.

\bibitem[\protect\citeauthoryear{Nvidia}{2022}]{nvidiaNVIDIAA100}
Nvidia.
\newblock {N}{V}{I}{D}{I}{A} {A}100 {G}{P}{U}s {P}ower the {M}odern {D}ata {C}enter --- nvidia.com.
\newblock \url{https://www.nvidia.com/en-us/data-center/a100/}, 2022.
\newblock [Accessed 17-01-2024].

\bibitem[\protect\citeauthoryear{Panda \bgroup \em et al.\egroup }{2022}]{panda2022sparsefed}
Ashwinee Panda, Saeed Mahloujifar, Arjun~Nitin Bhagoji, Supriyo Chakraborty, and Prateek Mittal.
\newblock Sparsefed: Mitigating model poisoning attacks in federated learning with sparsification.
\newblock In {\em International Conference on Artificial Intelligence and Statistics}, pages 7587--7624. PMLR, 2022.

\bibitem[\protect\citeauthoryear{Pillutla \bgroup \em et al.\egroup }{2022}]{pillutla2022robust}
Krishna Pillutla, Sham~M Kakade, and Zaid Harchaoui.
\newblock Robust aggregation for federated learning.
\newblock {\em IEEE Transactions on Signal Processing}, 70:1142--1154, 2022.

\bibitem[\protect\citeauthoryear{Rauniyar \bgroup \em et al.\egroup }{2023}]{rauniyar2023federated}
Ashish Rauniyar, Desta~Haileselassie Hagos, Debesh Jha, Jan~Erik H{\aa}keg{\aa}rd, Ulas Bagci, Danda~B Rawat, and Vladimir Vlassov.
\newblock Federated learning for medical applications: A taxonomy, current trends, challenges, and future research directions.
\newblock {\em IEEE Internet of Things Journal}, 2023.

\bibitem[\protect\citeauthoryear{Rodríguez-Gálvez \bgroup \em et al.\egroup }{2021}]{rodriguez2021enforcing}
Borja Rodríguez-Gálvez, Filip Granqvist, Rogier van Dalen, and Matt Seigel.
\newblock Enforcing fairness in private federated learning via the modified method of differential multipliers.
\newblock In {\em NeurIPS Workshop}, 2021.

\bibitem[\protect\citeauthoryear{Sandler \bgroup \em et al.\egroup }{2018}]{sandler2018mobilenetv2}
Mark Sandler, Andrew Howard, Menglong Zhu, Andrey Zhmoginov, and Liang-Chieh Chen.
\newblock Mobilenetv2: Inverted residuals and linear bottlenecks.
\newblock In {\em Proceedings of the IEEE conference on computer vision and pattern recognition}, pages 4510--4520, 2018.

\bibitem[\protect\citeauthoryear{Shejwalkar and Houmansadr}{2021}]{shejwalkar2021manipulating}
Virat Shejwalkar and Amir Houmansadr.
\newblock Manipulating the byzantine: Optimizing model poisoning attacks and defenses for federated learning.
\newblock In {\em NDSS}, 2021.

\bibitem[\protect\citeauthoryear{Shejwalkar \bgroup \em et al.\egroup }{2022}]{shejwalkar2022back}
Virat Shejwalkar, Amir Houmansadr, Peter Kairouz, and Daniel Ramage.
\newblock Back to the drawing board: A critical evaluation of poisoning attacks on production federated learning.
\newblock In {\em 2022 IEEE Symposium on Security and Privacy (SP)}, pages 1354--1371. IEEE, 2022.

\bibitem[\protect\citeauthoryear{Solans \bgroup \em et al.\egroup }{2021}]{solans2021poisoning}
David Solans, Battista Biggio, and Carlos Castillo.
\newblock Poisoning attacks on algorithmic fairness.
\newblock In {\em Joint European Conference on Machine Learning and Knowledge Discovery in Databases}, pages 162--177. Springer, 2021.

\bibitem[\protect\citeauthoryear{Vuolo and Levy}{2013}]{vuolo2013disparate}
Matthew~S Vuolo and Norma~B Levy.
\newblock Disparate impact doctrine in fair housing.
\newblock {\em New York Law Journal}, 2013.

\bibitem[\protect\citeauthoryear{Wang \bgroup \em et al.\egroup }{2022}]{wang2022understanding}
Jialu Wang, Xin~Eric Wang, and Yang Liu.
\newblock Understanding instance-level impact of fairness constraints.
\newblock In {\em International Conference on Machine Learning}, pages 23114--23130. PMLR, 2022.

\bibitem[\protect\citeauthoryear{Xie \bgroup \em et al.\egroup }{2020}]{xie2020dba}
Chulin Xie, Keli Huang, Pin~Yu Chen, and Bo~Li.
\newblock Dba: Distributed backdoor attacks against federated learning.
\newblock In {\em 8th International Conference on Learning Representations, ICLR 2020}, 2020.

\bibitem[\protect\citeauthoryear{Yin \bgroup \em et al.\egroup }{2018}]{yin2018byzantine}
Dong Yin, Yudong Chen, Ramchandran Kannan, and Peter Bartlett.
\newblock Byzantine-robust distributed learning: Towards optimal statistical rates.
\newblock In {\em International Conference on Machine Learning}, pages 5650--5659. PMLR, 2018.

\bibitem[\protect\citeauthoryear{Yue \bgroup \em et al.\egroup }{2023}]{yue2021gifair}
Xubo Yue, Maher Nouiehed, and Raed Al~Kontar.
\newblock Gifair-fl: A framework for group and individual fairness in federated learning.
\newblock {\em INFORMS Journal on Data Science}, 2(1):10--23, 2023.

\bibitem[\protect\citeauthoryear{Zeng \bgroup \em et al.\egroup }{2022}]{zeng2021improving}
Yuchen Zeng, Hongxu Chen, and Kangwook Lee.
\newblock Improving fairness via federated learning.
\newblock {\em arXiv preprint arXiv:2110.15545}, 2022.

\bibitem[\protect\citeauthoryear{Zhang \bgroup \em et al.\egroup }{2017}]{zhifei2017cvpr}
Zhifei Zhang, Yang Song, and Hairong Qi.
\newblock Age progression/regression by conditional adversarial autoencoder.
\newblock In {\em IEEE Conference on Computer Vision and Pattern Recognition (CVPR)}. IEEE, 2017.

\bibitem[\protect\citeauthoryear{Zhang \bgroup \em et al.\egroup }{2020}]{zhang2020fairfl}
Daniel~Yue Zhang, Ziyi Kou, and Dong Wang.
\newblock Fairfl: A fair federated learning approach to reducing demographic bias in privacy-sensitive classification models.
\newblock In {\em 2020 IEEE International Conference on Big Data (Big Data)}, pages 1051--1060. IEEE, 2020.

\bibitem[\protect\citeauthoryear{Zhang \bgroup \em et al.\egroup }{2022}]{zhang2022fldetector}
Zaixi Zhang, Xiaoyu Cao, Jinyuan Jia, and Neil~Zhenqiang Gong.
\newblock Fldetector: Defending federated learning against model poisoning attacks via detecting malicious clients.
\newblock In {\em Proceedings of the 28th ACM SIGKDD Conference on Knowledge Discovery and Data Mining}, pages 2545--2555, 2022.

\bibitem[\protect\citeauthoryear{Zhong \bgroup \em et al.\egroup }{2022}]{zhang2020hierarchically}
Zhengyi Zhong, Weidong Bao, Ji~Wang, Xiaomin Zhu, and Xiongtao Zhang.
\newblock Flee: A hierarchical federated learning framework for distributed deep neural network over cloud, edge, and end device.
\newblock {\em ACM Transactions on Intelligent Systems and Technology (TIST)}, 13(5):1--24, 2022.

\bibitem[\protect\citeauthoryear{Zhou \bgroup \em et al.\egroup }{2021}]{zhou2021deep}
Xingchen Zhou, Ming Xu, Yiming Wu, and Ning Zheng.
\newblock Deep model poisoning attack on federated learning.
\newblock {\em Future Internet}, 13(3):73, 2021.

\end{thebibliography}

\maketitle

\section{ Pseudo-Code of \methodName~Algorithm}

The detailed steps performed on the server and clients in \methodName~are summarized in Algorithm~\ref{alg:cap}.

\begin{algorithm}[ht]
\caption{\methodName~Algorithm}\label{alg:cap}
\begin{algorithmic}
\STATE \textbf{Server:}

\STATE  {Initialize global model parameter $\theta_{g}^0$}

\FOR{each round $t$ = 0, 1, · · ·}
    \STATE Server randomly selects a subset of $n$ clients
    \STATE Server sends the latest global model $\theta_{g}^t$ to $n$ clients
    \FOR{each client $k$ = 0, 1, · · · $n$} 
        \STATE  {Gather local update ($\theta_{k}^{t}$)}
        \STATE  Calculate clients' aggregation weight $\hat{w}_{k}^{t}$ using the preferred aggregation method 
    \ENDFOR
\STATE {Update the global model parameter }
\STATE {$\theta_{g}^{t+1}=\theta_{g}^{t} + \sum^n_{k=1}\hat{w_{k}}\cdot(\theta^{t}_k - \theta^{t-1}_k)$}
\ENDFOR

\STATE {\textbf{Client (\textit{benign}):}}

\STATE {Receive global model parameter $\theta_{g}^t$ from the server;}
\FOR{$(x_{i},y_{i},g_{i})\in \mathcal{D}$} 
    \STATE {$ \underset{\theta^{t}_{g} \in \Theta}{\min} \mathcal{L}(f(x_i,\theta^{t}_{g}), y_i)$ \quad $\textup{s.t.} \, \phi(f,g_i) \leq \mu$}
\ENDFOR    
\STATE    {Send the updated $\theta^{t}_{g}$ to the server.}

\STATE {\textbf{Client (\textit{malicious}):}}

\STATE {Receive global model parameter $\theta_{g}^t$ from the server;}
\FOR{$(x_{i},y_{i},g_{i})\in \mathcal{D}$} 
    \STATE {$ \underset{\theta^{t}_{b} \in \Theta}{\min} \mathcal{L}(f(x_i,\theta^{t}_{b}), y_i)$ \quad $\textup{s.t.} \, \phi(f,g_i) \leq \mu$}
\ENDFOR    

\FOR{$(x_{i},y_{i},g_{i})$ and $(x_{j},y_{j},g_{j})\in \mathcal{D}$ and $ g_j=\tau$} 
   \STATE {infl$(j,i)= \int_{g_{j} = \tau} \Theta(x_i, x_j ; \theta_g^t) (\frac{\partial \mathcal{L}(w, y_i)}{\partial w}$}
  \STATE \qquad \qquad \qquad{$+ \frac{\partial \phi(f,g_i)}{\partial l}) \text{dPr}(x_j,y_j,g_j)$}
\ENDFOR  
\STATE Select biasing dataset $\mathcal{D}^{bias}$
\FOR{$(x_{i},y_{i},g_{i})\in \mathcal{D}^{bias}$} 
    \STATE{$\underset{\theta_p}{\min}  \frac{1}{|\mathcal{D}^{bias}|} \sum_{i=1}^{|\mathcal{D}^{bias}|} \mathcal{L}(l(x_i; \theta_p), y_i)$} 
    \STATE{$+ \gamma \sum_{\theta^*\in \theta_p} h(\Delta\theta^*)^2 + \rho ||\theta_p - \theta_b||_2$}
\ENDFOR
\STATE    {Send the updated $\theta^{t}_{g}$ to the server.}

\end{algorithmic}
\end{algorithm}
\vspace{-2mm}

\section{FL Settings}
We evaluate the proposed~\methodName~on three datasets: CelebA, Adult Income, and UTK Faces. Specifically, the CelebA dataset contains $200,288$ images from $9,343$ celebrities, and it is divided into training ($80\%$) and testing ($20\%$) sets with no overlapping celebrity ID. To emulate the cross-silo FL setup, the training dataset is further divided into $100$ silos, with each silo representing a client with no overlapping celebrity ID. The Adult Income dataset is distributed among $40$ clients, and a test set of $10\%$ of the data is kept for evaluating the global model. The UTK Faces dataset is divided into $90\%$ training and $10\%$ testing data. The training dataset is randomly distributed into $20$ clients, each of which in this setup represents a different device or location. 

\section{EXPERIMENTAL DETAILS}
\subsection{Model Description}
We evaluate \methodName~on three datasets, namely, CelebA, Adult Income, and UTK Faces.
The details of the FL model used for each dataset are presented below:

\begin{enumerate}[label={(\arabic*)}]
\item \textit{CelebA}: We train a ResNet34~\cite{he2016deep} model with a single fully connected layer, equipped with a sigmoid activation function for binary classification (i.e., classifying whether the celebrity in the image is smiling or not). The input layer is modified to accommodate the image dimensions of 128x128 pixels, characteristic of the CelebA dataset to reduce the computation cost at the clients. We use Adam optimizer with a learning rate of $0.01$ with $10\%$ decay rate.

\item \textit{Adult Income}: We use a Multi-layer Perceptron (MLP) model containing two hidden layers, each with 32 and 16 units, respectively, and a single unit output layer for binary classification (i.e., classifying whether a subject’s income is above 50k or not). We use SGD with a learning rate of $0.01$ and momentum of $0.9$

\item \textit{UTK Faces}:
We train a MobileNetV2~\cite{sandler2018mobilenetv2} model with an input size of $128$x$128$. We use Adam optimizer with a learning rate of $0.05$ with $10\%$ decay rate.
\end{enumerate}

\vspace{-1mm}
\subsection{Evaluation Settings}
\vspace{-1mm}
In our evaluation, we adjust the FL setup in each dataset to accommodate data from different genres. 
\begin{enumerate}[label={(\arabic*)}]

\item \textit{CelebA}: We use $100$ clients to train the model for $400$ communication rounds with a client participation rate of $0.3$, a $\gamma$ value of $0.5$ and $\rho$ value of $0.7$. For the biasing dataset, we use $\kappa$ of $0.4$ to achieve highest fairness and utility trade-off. 

\item \textit{Adult Income}: We use $40$ clients to train the model for $100$ communication rounds with a client participation rate of $0.4$, a $\gamma$ value of $0.4$ and $\rho$ value of $0.7$.

\item \textit{UTK Faces}: We use $20$ clients to train the model for $200$ communication rounds with a client participation rate of $0.6$ , a $\gamma$ value of $0.5$ and $\rho$ value of $0.6$. 
\end{enumerate}

We use $\mu$ of $0.8$ for all the datasets as it ensures that the local fairness constraints tries to achieve $80\%$ rule for fairness~\cite{biddle2017adverse,vuolo2013disparate}. We perform experiments under non-IID scenarios. We distribute the dataset among the clients such that the distribution of sensitive attributes (G) is non-IID and can be controlled using a heterogeneity controller $\alpha$, where $\alpha \rightarrow \infty$ corresponds to IID distributions. For the experiments unless mentioned otherwise, we use $\alpha$ value of $0.1$. 

\subsection{Secure Aggregation Methods}
We assess the stealth of the proposed~\methodName~attack aginst robust aggregation methods. Four secure federated learning aggregation algorithms, namely SparseFed~\cite{panda2022sparsefed}, Krum~\cite{yin2018byzantine}, LoMAR~\cite{li2021lomar}, and FLDetector~\cite{zhang2022fldetector}, are considered in our evaluation. These methods rely on model similarity or weight distance metrics for secure aggregation.

Krum is a Byzantine-resilient algorithm that selects only one client per round and rejects outlier model updates. SparseFed, on the other hand, employs an \(L_2\) threshold to discard parameters that exceed a certain threshold. LoMar assesses the maliciousness of client updates by analyzing the statistical features of model parameters in relation to their neighbors, utilizing a non-parametric relative kernel density estimation approach. FLDetector is an unsupervised method designed to detect malicious clients by evaluating the consistency between received and predicted model updates from clients.

\begin{table*}[t]
\vspace{-10pt}
\caption{Effect of the size of biasing dataset on the performance of~\methodName} \label{table:bias_dataset}
\begin{center}
\resizebox{0.75\linewidth}{!}{
\begin{tabular}{@{}cccccccccc@{}}
\toprule
         \multirow{4}{6em}{\centering\textbf{$\kappa$}} &\multicolumn{3}{c}{\textbf{CelebA} ($\epsilon=0.1$)} &\multicolumn{3}{c}{\textbf{Adult Income} ($\epsilon=0.2$)} &\multicolumn{3}{c}{\textbf{UTK Faces} ($\epsilon=0.2$)}\\ \cmidrule(r){2-4} \cmidrule(r){5-7} \cmidrule(r){8-10}
         
        &{\textbf{EOD} ($\downarrow$)} &{\textbf{DPD}($\downarrow$)} &\multirow{2}{2.5em}{\textbf{Utility}} &{\textbf{EOD} ($\downarrow$)} &{\textbf{DPD}($\downarrow$)} &\multirow{2}{2.5em} {\textbf{Utility}} &{\textbf{EOD} ($\downarrow$)} &{\textbf{DPD}($\downarrow$)} &\multirow{2}{2.5em}{\textbf{Utility}} \\ \cmidrule(r){2-3}   \cmidrule(r){5-6} \cmidrule(r){8-9}     
    
        &\textbf{Gender} &\textbf{Gender} &\Gape[1ex][1ex]{($\uparrow$)} &\textbf{Race} &\textbf{Race} & \Gape[1ex][1ex]{($\uparrow$)} &\textbf{Race} & \textbf{Race} &\Gape[1ex][1ex]{($\uparrow$)}\\\midrule

0.1 &0.25 &0.27 &89\% &0.29 &0.30 &82\% &0.27 &0.25 &85\%\\
                    \midrule
     
0.2 &0.32 &0.34 &88\% &0.32 &0.35 &81\% &0.29 &0.28 &85\%\\
                   \midrule
     
0.3 &0.35 &0.38 &88\% &0.37 &0.39 &80\% &0.30 &0.29 &84\%\\
                      \midrule
     
0.4 &0.41 &0.43 &88\% &0.41 &0.44 &80\% &0.37 &0.34 &84\%\\
                            \midrule
0.5 &0.47 &0.48 &76\% &0.45 &0.44 &71\% &0.43 &0.44 &70\%\\
                            \midrule
                            
0.6 &0.47 &0.49 &71\% &0.47 &0.45 &70\% &0.44 &0.45 &63\%\\
                          \bottomrule
\end{tabular}
}
\end{center}
\end{table*}
\section{Additional Evaluation}
\subsection{Impact of the Biasing Dataset Size}
In our model poisoning fairness attack, the size of the biasing dataset is a critical factor. We conducted experiments to understand the impact of various biasing dataset sizes.
These sizes are represented by the fraction $\kappa$, which indicates the proportion of data originating from the privileged group.
It's important to note that the biasing dataset cannot be too large.
It should not match the full size of the dataset from the privileged group. A biasing dataset of this magnitude could lead to overfitting, compromising the model's ability to generalize effectively.

Additionally, when $\kappa$ is substantial, it may not be possible to exclusively select samples with negative influence scores. In such scenarios, it becomes necessary to include samples with small positive influence scores. The results of our experiments, detailed in Table~\ref{table:bias_dataset}, reveal a clear trend: both Equal Opportunity Difference (EOD) and Difference in Positive Predictions (DPD) increase as the size of the biasing dataset grows. However, when $\kappa$ exceeds $0.4$, the model's accuracy significantly declines. This decline is attributed to the instability caused by the poison neurons introduced during adversarial training, leading to an overfitting of the model. This overfitting, in turn, adversely affects the model's accuracy, highlighting the need for a balanced biasing dataset size to maintain model effectiveness while executing the attack.

\subsection{Impact of Different FL Data Settings}

To evaluate the effectiveness of \methodName under different levels of data heterogeneity, we conducted experiments considering different data distributions, including both IID and non-IID scenarios. The dataset was distributed among $n$ clients in a way that the distribution of sensitive attributes $(G)$ becomes non-IID and can be controlled using a heterogeneity parameter $\alpha$, where $\alpha \rightarrow \infty$ corresponds to IID distributions. To introduce this heterogeneity, we utilized a power law distribution~\cite{clauset2009power}.

The results in Table~\ref{table:iid} demonstrate that data heterogeneity does have a notable impact on the attack effectiness. Non-IID data distribution is relatively easier to attack due to its general impact on fairness. However, it's worth noting that even in scenarios with an IID data distribution, \methodName~ can still significantly affect fairness, showcasing its effectiveness under various conditions.

\begin{table}[t]
\vspace{-3mm}
\caption {Performance under different FL  data settings.}\label{table:iid}
\vspace{-5mm}
\begin{center}
\resizebox{0.96\linewidth}{!}{
\renewcommand{\arraystretch}{1.2}
\begin{tabular}{|c|c|c|c|c|c|c|c|}
    \hline
    \multirow{3}{5em}{\centering\textbf{Metric}} &\multirow{3}{6em}{\centering\textbf{Attack}} &\multicolumn{3}{c|}{\textbf{CelebA}} &\multicolumn{3}{c|}{\textbf{UTK Faces}}\\\cline{3-8}
    &&\multicolumn{3}{c|}{\textbf{$\alpha$}}&\multicolumn{3}{c|}{\textbf{$\alpha$}}\\\cline{3-8}
    &&\textbf{0.1} &\textbf{10} &\textbf{100}&\textbf{0.1} &\textbf{10} &\textbf{100}\\

    \hline
    \multirow{3}{5em}{\centering \textbf{EOD ($\downarrow$)}}
    &Gradient-based &0.25 &0.24 &0.16 &0.29 &0.25 &0.16\\
    &Anchoring-based &0.24 &0.23 &0.18 &0.27 &0.23 &0.16\\
    
    &\textbf{\methodName} &\textbf{0.41} &\textbf{0.39} &\textbf{0.35} &\textbf{0.37} &\textbf{0.32} &\textbf{0.31}\\
    \hline
    \multirow{3}{5em}{\centering\textbf{DPD ($\downarrow$)}}
    &Gradient-based &0.23 &0.18 &0.17 &0.31 &0.18 &0.16\\
    &Anchoring-based &0.22 &0.21 &0.18 &0.28 &0.19 &0.16\\
    &\textbf{\methodName} &\textbf{0.43} &\textbf{0.40} &\textbf{0.34} &\textbf{0.34} &\textbf{0.30} &\textbf{0.28}\\
    \hline
    \multirow{3}{5em}{\centering\textbf{Utility ($\uparrow$)}}
    &Gradient-based &85\% &87\% &88\% &82\% &84\% &84\%\\
    &Anchoring-based &81\% &83\% &84\% &81\% &81\% &83\%\\
    &\textbf{\methodName} &88\% &89\% &90\% &84\% &86\% &86\%\\
    \hline

\end{tabular}
}
\end{center}
\end{table}

\subsection{Real World Case Study}

In order to illustrate the real-world implications of fairness attacks, we perform \methodName on a movie recommendation system. The target system utilizes Neural Collaborative Filtering (NCF)~\cite{he2017neural}, a popular algorithm for recommending movies based on user preferences or the preferences of similar users. The recommendation system is trained in a federated setting using the MovieLens 1M dataset~\cite{harper2015movielens}. The objective of this attack is to bias the model in such a way that all users receive recommendations for a specific movie, regardless of whether it aligns with their individual preferences. Let $E$ be the embedding matrix of the movies, where each row represents the embedding of a movie, and let $b$ be the bias vector of size $M$ (number of movies) representing the bias term added to each movie's embedding. Let $R$ be the set of ratings, where each rating $(u, i, r)$ represents the rating $r$ given by user $u$ to movie $i$. 

The attacker's objective is to maximize the bias term for a targeted movie $i_t$ while minimizing the bias terms for all other movies $i \neq i_t$. The attacker also wants to ensure that the modified model still performs well on the training data, represented by the mean squared error (MSE) loss.
Formally, the attacker's optimization problem can be written as:

\begin{equation} 
\setlength{\abovedisplayskip}{3pt}
\setlength{\belowdisplayskip}{3pt}
\begin{split}
  & {\max} \quad b[i_t] - \lambda \sum_{i \neq i_t} b[i]\\
  &\textup{s.t.}\quad \frac{1}{N} \sum_{(u,i,r)\in R} (r - (E[i] + b[i])^T E[u])^2 \leq \epsilon,\\
  &\quad \quad  -c \leq b[i]\leq c, \forall i,
\end{split}
\label{eq4}
\end{equation}
where $\lambda$ is a hyperparameter controlling the tradeoff between maximizing the bias for the targeted movie and minimizing the bias for other movies, $N$ is the number of ratings in $R$, and $\epsilon$ is the allowed error in the MSE loss. The second constraint ensures that the bias terms remain within a certain range, represented by the constant $c$. In this study, we conducted experiments using the MovieLens dataset, which was divided into $90\%$ training data and $10\%$ testing data. The training data was distributed among 500 clients, each with a participation probability of $0.2$. Our targeted movie for the fairness attack was "Mortal Kombat (1995)," which had an average rating of $2.78$.

\begin{table}
\vspace{-10pt}
\caption {Performance of~\methodName against movie recommendation system} \label{table-10} 
\vspace{-2mm}
\begin{center}
\resizebox{0.8\linewidth}{!}{
\begin{tabular}{@{}cccc@{}}
\toprule
         \textbf{Method}  &\textbf{Attack Scenario} &\textbf{Prob-T} &\textbf{Utility (RMSE)}\\ \midrule
\multirow{3}{3em}{\textbf{FedAvg}} &No Attack &0.32 & 0.08\\ 
                                    &\methodName($\epsilon=0.2$) &0.57 & 0.11\\ 
                                    &\methodName($\epsilon=0.3$) &0.84 & 0.23\\
\bottomrule
\multirow{3}{3em}{\textbf{q-FFL}} &No Attack &0.34 & 0.10\\
                                       &\methodName($\epsilon=0.2$) &0.55 & 0.10\\ 
                                       &\methodName($\epsilon=0.3$) &0.81 & 0.22\\
                                       \bottomrule
\end{tabular}
}
\vspace{-4mm}
\end{center}
\end{table}

Table~\ref{table-10} shows the results of our attack against the movie recommendation system. Moreover, we only adopted FedAvg aggregation and q-FFL methods in FL as other fairness optimization strategies, such as~\fedfb and~\fairfed, require the dataset to have demographic information, which is unavailable for the \textit{MovieLens 1M} dataset. Since the objective of the fairness attack in this scenario is to introduce bias into the model towards a specific sample rather than exacerbating unfairness among different groups, the EOD and DPD fairness metrics are not used. Instead, in Table~\ref{table-10}, we provide the probability (Prob-T) of the targeted movie being recommended to any user, while evaluating the utility using RMSE (Root Mean Square Error) loss, which measures the accuracy of the model's rating predictions for each user. Specifically, the results show that the targeted movie has a probability of $0.32$ of being recommended without any attack. Under attack, this probability increases significantly to $0.84$. Even with a malicious participant probability of $\epsilon=0.2$, the model's fairness was disrupted, resulting in a recommendation probability of $0.57$ for the targeted movie. The results also show that the inclusion of q-FFL-based aggregation has minimal impact on both the recommendation probability and the utility.





\end{document}